\documentclass[sn-mathphys,Numbered]{sn-jnl}

\usepackage{graphicx}%
\usepackage{multirow}%
\usepackage{amsmath,amssymb,amsfonts}%
\usepackage{amsthm}%
\usepackage{mathrsfs}%
\usepackage[title]{appendix}%
\usepackage{textcomp}%
\usepackage{manyfoot}%
\usepackage{booktabs}%
\usepackage{algorithm}%
\usepackage{algorithmicx}%
\usepackage{algpseudocode}%
\usepackage{listings}%

\usepackage{lipsum}
\usepackage[printonlyused, nolist]{acronym}
\usepackage{soul} 
\usepackage{caption}    
\usepackage{subcaption} 
\usepackage[table]{xcolor}
\usepackage{booktabs}   
\usepackage{gensymb}    
\usepackage{siunitx}
\usepackage{dcolumn}
\usepackage[square,numbers]{natbib}
\usepackage{changes}%

\newcolumntype{C}[1]{>{\centering\arraybackslash}p{#1}}
\newcolumntype{d}[1]{D{.}{.}{#1}}
\definecolor{colortrain}{HTML}{DAE8FC}
\definecolor{colortest}{HTML}{F8CECC}
\definecolor{colornotperformed}{HTML}{F5F5F5}
\definecolor{colornottrained}{HTML}{BBE3BA}

\newcommand{\mc}[1]{\multicolumn{1}{c}{#1}}
\begin{document}
\begin{acronym}
    \acro{NN}{neural network}
    \acro{RMSE}{root mean squared error}
    \acro{ABTC}{area between the thickness curves}
    \acro{DL}{deep learning}
    \acro{GNN}{graph neural network}
    \acro{FEM}{finite element method}
\end{acronym}

\title{Predicting Wall Thickness Changes in Cold Forging Processes: An Integrated FEM and Neural Network approach$^1$}

\author[1]{\fnm{Sasa} \sur{Ilic}}\email{ilic.sasa@fh-swf.de}
\equalcont{These authors contributed equally to this work.}

\author[1]{\fnm{Abdulkerim} \sur{Karaman}}\email{karaman.abdulkerim@fh-swf.de}
\equalcont{These authors contributed equally to this work.}

\author[2]{\fnm{Johannes} \sur{Pöppelbaum}}\email{poeppelbaum.johannes@fh-swf.de}
\equalcont{These authors contributed equally to this work.}

\author[2]{\fnm{Jan Niclas} \sur{Reimann}}\email{reimann.janniclas@fh-swf.de}
\equalcont{These authors contributed equally to this work.}

\author[1]{\fnm{Michael} \sur{Marré}}\email{marre.michael@fh-swf.de}
\equalcont{These authors contributed equally to this work.}

\author[2]{\fnm{Andreas} \sur{Schwung}}\email{schwung.andreas@fh-swf.de}
\equalcont{These authors contributed equally to this work.}

\affil*[1]{\orgdiv{Labor für Massivumformung}, \orgname{South Westphalia University of Applied Sciences}, \orgaddress{\street{Frauenstuhlweg 31}, \postcode{58644} \city{Iserlohn},  \country{Germany}}}

\affil[2]{\orgdiv{Department for Automation Technology and Learning Systems}, \orgname{South Westphalia University of Applied Sciences}, \orgaddress{\street{Lübecker Ring 2}, \postcode{59494} \city{Soest}, \country{Germany}}}

\footnotetext[1]{This preprint is currently under review at the \textit{Journal of Intelligent Manufacturing}.}

\abstract{This study presents a novel approach for predicting wall thickness changes in tubes during the nosing process. Specifically, we first provide a thorough analysis of nosing processes and the influencing parameters. We further set-up a Finite Element Method (FEM) simulation to better analyse the effects of varying process parameters. As however traditional FEM simulations, while accurate, are time-consuming and computationally intensive, which renders them inapplicable for real-time application, we present a novel modeling framework based on specifically designed graph neural networks as surrogate models. To this end, we extend the neural network architecture by directly incorporating information about the nosing process by adding different types of edges and their corresponding encoders to model object interactions. This augmentation enhances model accuracy and opens the possibility for employing precise surrogate models within closed-loop production processes. The proposed approach is evaluated using a new evaluation metric termed area between thickness curves (ABTC). The results demonstrate promising performance and highlight the potential of neural networks as surrogate models in predicting wall thickness changes during nosing forging processes.}


\maketitle
\section{Introduction}
Tubes are essential components of the drivetrain of millions of cars, lorries and other vehicles. In addition to their use in the powertrain, tubes are also used in steering racks, seat structures and various other automotive and manufacturing components. These components are made from tubes using various forging processes. Nosing, a form of cold forging, is used to shape these tubes by reducing their diameter.

The nosing process, in which the diameter of a tube is reduced, results in significant changes to the wall thickness. As the tube is formed and its diameter decreases, material is displaced, leading to an increase in wall thickness. The wall thickness, which refers to the distance between the inner and outer surfaces of the tube, plays a crucial role in the mechanical properties of the component, including its strength, rigidity, and ability to resist deformation. A consistent and well-defined wall thickness ensures that the tube can maintain its structural integrity during use.

The wall thickness change during the nosing process, however, is a key factor to control. As the outer diameter of the tube is reduced, the wall thickness increases due to material flow. This variation is essential to monitor, as it affects the dimensional accuracy and overall quality of the final product. Inconsistent wall thickness changes can lead to structural weaknesses, reducing the stability and performance of the part. Therefore, accurate control and prediction of wall thickness changes are vital for ensuring that the formed components meet the required specifications and maintain high quality in production. 

This work deals with theoretical and practical aspects of the nosing process. Firstly, a comprehensive analysis of industrial nosing processes is provided, detailing the underlying physical relationships and defining the key influencing parameters. This analysis not only unifies existing results from the literature, but also presents new insights into the mechanics of wall thickness variations during nosing processes, as this foundational understanding is crucial for developing more accurate predictive models.

Building on this detailed analysis, a FEM model is developed to simulate the nosing process. The FEM model is designed to capture the complex deformation behaviors of tubes during nosing, considering various parameters such as  friction conditions or forming degrees. This model provides a robust framework for generating simulation data with high degree of accuracy, which is essential for training the neural network model.

\begin{figure}[h!]
\centering
\includegraphics[width=1\textwidth]{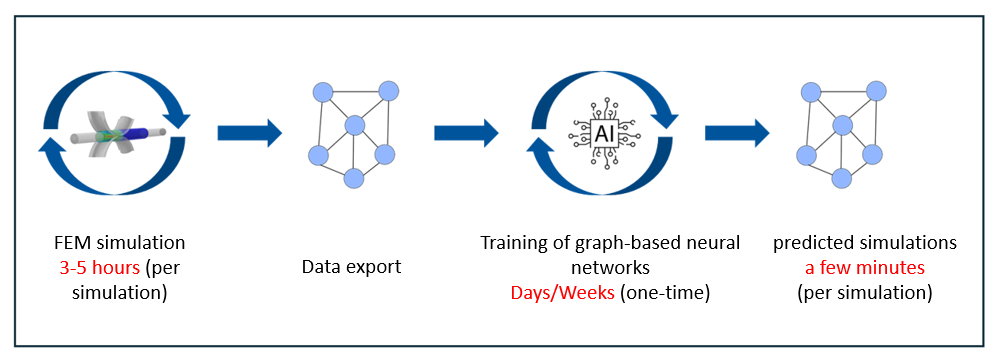}
\caption{Approach for creating the AI Model.}
\label{fig:my_label1}
\end{figure}

Recognizing the computational limitations of traditional FEM simulations, especially for real-time applications, a novel neural network architecture tailored to the nosing process is introduced (see \autoref{fig:my_label1}.). This approach leverages graph neural networks (GNNs) due to their ability to effectively model the interactions within the nosing process. By incorporating different parameters and boundaries, the GNN architecture accurately captures the dynamic relationships between various elements of the forging process. This results in a surrogate model that is computationally efficient in deployment and capable of real-time predictions, maintaining a level of accuracy comparable to FEM simulations.

In addition to developing the neural network model, a new evaluation metric, the area between thickness curves (ABTC), is proposed. This metric provides a quantitative measure of the differences in wall thickness changes predicted by various models, offering a more nuanced and precise evaluation of model performance.

The approach is validated through extensive experiments, demonstrating the promising performance of the neural network model. The results highlight the potential of neural networks as surrogate models for FEM simulations, showcasing their ability to predict wall thickness changes with high accuracy and reduced computational costs. The experiments include several scenarios to validate the strengths of the approach, ensuring that the proposed model is robust and generalizable across different scenarios.

Overall, this research represents a significant advancement in the field of cold forging processes. By integrating detailed FEM simulations with advanced neural network techniques, a powerful tool for predicting wall thickness changes in nosing processes is provided. This has the potential to significantly enhance the efficiency and accuracy of manufacturing operations, leading to cost savings and improved product quality. The work lays the foundation for further exploration and application of machine learning in forging, paving the way for more intelligent and adaptive manufacturing systems.

The paper is structured as follows: \autoref{sec:related_work} discusses the related work and \autoref{sec:process} introduces the forging process. This is followed by the \ac{NN} architecture to approximate FEM simulations in \autoref{sec:architecture} and our obtained results in \autoref{sec:results}. \autoref{sec:conclusion} concludes the paper.

\subsection{Related work} \label{sec:related_work}
In the following, we present related work on the analysis of nosing processes as well as surrogate models to replace FEM simulations.

\subsubsection{Nosing Processes}
In the field of forging techniques, the large number of parameters has a significant influence on forging processes \cite{Karaman2023InvestigationOF}. This is also the case in the nosing process. The process of nosing tubes is relatively underrepresented in scientific research. There is a lack of in-depth studies that delve into the characteristic features and limitations of this method. Exceptions include some works conducted for thin-walled tubes \cite{Ebe80} and thick-walled tubes \cite{Haa83}. The description of the nosing process in  literature often follows a standardized format. In the process, the stamping force pushes the tube into the conical part of the die, causing the raw part to proceed towards the die. The outer diameter is reduced upon exiting the die. The inlet and outlet radius form freely. The reduced diameter is rounded to facilitate material flow \cite{Ebe80}. During nosing, there is uncontrolled thickening of the wall as the material is displaced both radially and tangentially. This results in variable wall thicknesses along the length of the tube \cite{Alb90}. The production parameters varied during nosing have a strong influence on the change in wall thickness. These influences are adjusted in production through worker experience. However, the increase in wall thickness during such nosing processes is difficult to predict and depends on a variety of parameters (material properties, friction conditions, forming degrees, etc.). Due to the challenging flow behavior of the material towards wall thickness increase, the tubes are designed with larger dimensions for safety. As a result of nosing, the tube ends become too long and unevenly formed, requiring them to be cut straight. This leads to a higher scrap rate, with potential savings of up to 8\% of the tube dimension. Furthermore, this potential for savings would also affect applications in the field of flow forming, for example, in the flow forming of cup and sleeve-shaped flow-formed parts, whose number far exceeds that of hollow shafts in the automotive sector. Hence, a better and more general understanding of the influencing factors and parameters during the production process is required. To understand the influence of parameters in difficult to model processes, often FEM simulation are employed. FEM simulations are conducted to adjust the preform geometry so that uniform wall thickness can be achieved in the final component \cite{TANG1982293}. Tracing backward methods are also performed to optimize the preform using historical approaches \cite{HWANG19871}, or with convolutional neural networks (CNNs) \cite{KIM20244625}.. Furthermore, FEM simulations are conducted to determine process limits such as buckling behavior \cite{FATNASSI1985643}. Investigations are also conducted on the influences of various process parameters, such as tube length, tube thickness, friction, or die opening angle, in eccentrically deformed tubes \cite{KWAN2003530}, as well as to form failure models \cite{articlelfm10}. It can therefore be stated that FEM analyses provide good prediction results, but are time-consuming and cannot capture changes during the forging process. For this reason, AI is increasingly being used in manufacturing and metal forging. In this study, the application of machine learning, and closed-loop control for real-time control of bending processes in bending to compensate for springback was investigated \cite{MOLITOR23}. \cite{SAKARIDIS2022104240} developed a machine learning model to predict the crash response of tubular structures, aiming to combine the efficiency of analytical methods with the accuracy of finite element simulations and accurately model the complex nonlinear crushing responses. Recent advances in machine learning, particularly in \ac{DL}, have opened up new avenues for creating surrogate models that can approximate the results of FEM simulations with significantly reduced computational costs. By leveraging the strengths of \ac{GNN}s and physics-informed neural networks (PINNs), models can be developed mimic the physical behavior of materials. An example of this is the application of a machine learning method to investigate buckling behavior in tube structures, aiming to better predict these phenomena \cite{SAKARIDIS2022104240}. With the continuous advancements in AI \ac{DL}, their applications in manufacturing and forging processes are expected to grow, offering more efficient and accurate predictive capabilities.

\subsubsection{Finite Element Surrogate Models}
Applying machine learning (ML) techniques to analyze manufacturing processes is becoming a common tool - especially the automated analysis of surface defects or surface features. Already in 2004 Özel and Karpat \cite{OZEL2005467} used a vanilla feedforward neural network to predict the surface roughness and tool wear based on the rockwell-C hardness, the cutting speed, the feed rate, the axial cutting length and the forces in x, y, and z- direction respectively. In 2018 Pimenov et al. \cite{Pimenov2018} compared the performance of a similar neural network architecture, very small and basic multi-layer perceptron (MLP), with radial basis functions, random forest and tree search in a milling process. Nogueira et al. \cite{NOGUEIRA2022657} analyzed the performance of MLPs, as well as convolutional networks to classify the surface of turned surfaces and show its applicability. Reyes-Luna et al. \cite{REYESLUNA2023458} developed a surrogate model of a printing application by training a genetic algorithm to tune the parameters of their surrogate model. Kim et al. \cite{KIM20238228} use a basic three-layer MLP to directly predict process parameters. Daoud et al. \cite{DAOUD2021529} also used a basic three-layer MLP to predict residual stress fields after shot peening a steel surface achieved similar results to that of an FEM simulation. Gondo and Arai \cite{Gondo2022} predicted the thickness at three different spots of a workpiece manufactured by multi-pass spinning by using a basic three-layer MLP as well, while Mujtaba et al. \cite{Mujtaba2023} used a similar architecture to create a model to predict local thermal profiles during automated fibre placement (AFP). The tuned model was 4 times faster than the used FEM. Nabian and Meidani~\cite{DBLP:journals/corr/abs-1806-02957}, Haghighat et al.~\cite{HAGHIGHAT2021113741}, and Karumuri et al.~\cite{karumuri2020simulator} applied a basic multilayer perceptron and incorporated the partial differential equations (PDEs) of their underlying physical system as learning constraints. Kubik et al.~\cite{Kubik2022}\cite{Kubik2023} use approaches such as a multiclass support vector machines to predict the wear state in a blanking process or using a principal component analysis as a feature extractor in combination with small neural networks (one to threee layers) for wear detection in roll forming. Most of these approaches are founded on basic MLPs using scalars of material parameters or manufacturing parameters. However, MLPs are inherently limited for modeling FEM models as they have a rather general, non specialized architecture.
Within various architectures to solve different types of PDE the most promising approach provides the combination of graph neural networks (GNNs) with ideas from physics informed neural networks (PINNs). GNNs, which are also used for process scheduling (Hameed \& Schwung~\cite{hameed2023graph}) or event based systems (Lassoued \& Schwung~\cite{petriRL2024}), consist, analog to FEMs, of nodes and edges. Information between nodes is communicated via the encoded representation of features of all connected nodes and edges, where exactly this encoding process is trained in the learning process.
Chan et al.~\cite{CHAN20081170} use original feed forward neural networks (FFNNs) to predict stresses of a mechanical forming process of the simulated FEM mesh. Sanchez-Gonzalo et al.~\cite{pmlr-v119-sanchez-gonzalez20a} expanded this idea to GNNs and simulated fluid dynamics of particles (sand, water, ...)  in a closed container under varying conditions (dropping, shaking, ramps, ...). They also experimentally showed the performance gain of using GNNs compared to continuous convolutions~\cite{ummenhofer2019lagrangian}. Gulakala et al.~\cite{gulakala2022} accelerated FEM simulations by outsourcing the boundary value problem of FEM simulators to GNNs. They used embedded node- and edge information of the FEM simulator and process decoded the GNN prediction to predict von-Mises stress distributions after deformations of an object. Shivaditya et al.~\cite{shivaditya2022graph} use a 1-step GNN predictor to predict the die of a yoke metal forging process to measure the wear depending on initial conditions such as temperature of the object and friction between the billet and the deformable die. Deshpande et al.\cite{deshpande2023magnet} extend the ideas of the well known U-Net architecture~\cite{unet} and the develop an innovative Multichannel Aggregation (MAg) layer. \cite{Pfaff2021} proposed an efficient way to incorporate dynamic edges by splitting up the problem space into a world space and a mesh space. Meshes from the world space get transformed into mesh space to model contact information between parts. The manipulated mesh gets encoded, all node and mesh embeddings gets updated, and the 1-step difference gets extracted by a decoder and passed to an integrator.
In conclusion, there exist a multitude of approaches to accelerate simulations using machine learning models as surrogate models since these high-fidelity simulations are too slow for practical use in any closed-loop production process. However, either the presented architectures are rather generic and used for benchmark examples only without an application context, or application near approaches mostly rely on MLPs with their inherent limitations. In this work, we propose a novel GNN-based network specifically taylored for the use in nosing processes. Particularly, using the GNN architecture from\cite{Pfaff2021} as a baseline, we incorporate specialized encoders to model the various forms of interactions between different objects to be considered in nosing processes and propose a novel evaluation metric for GNN model performance.

\section{Analysis of Nosing Processes} \label{sec:process}
In this section, we present a thorough discussion and analysis of the paramters influencing the nosing process. Subsequently, we present a detailed FEM-simulation of such processes. 

\subsection{Description of the nosing process}
\begin{figure}[h!]
\centering
\includegraphics[width=1\textwidth]{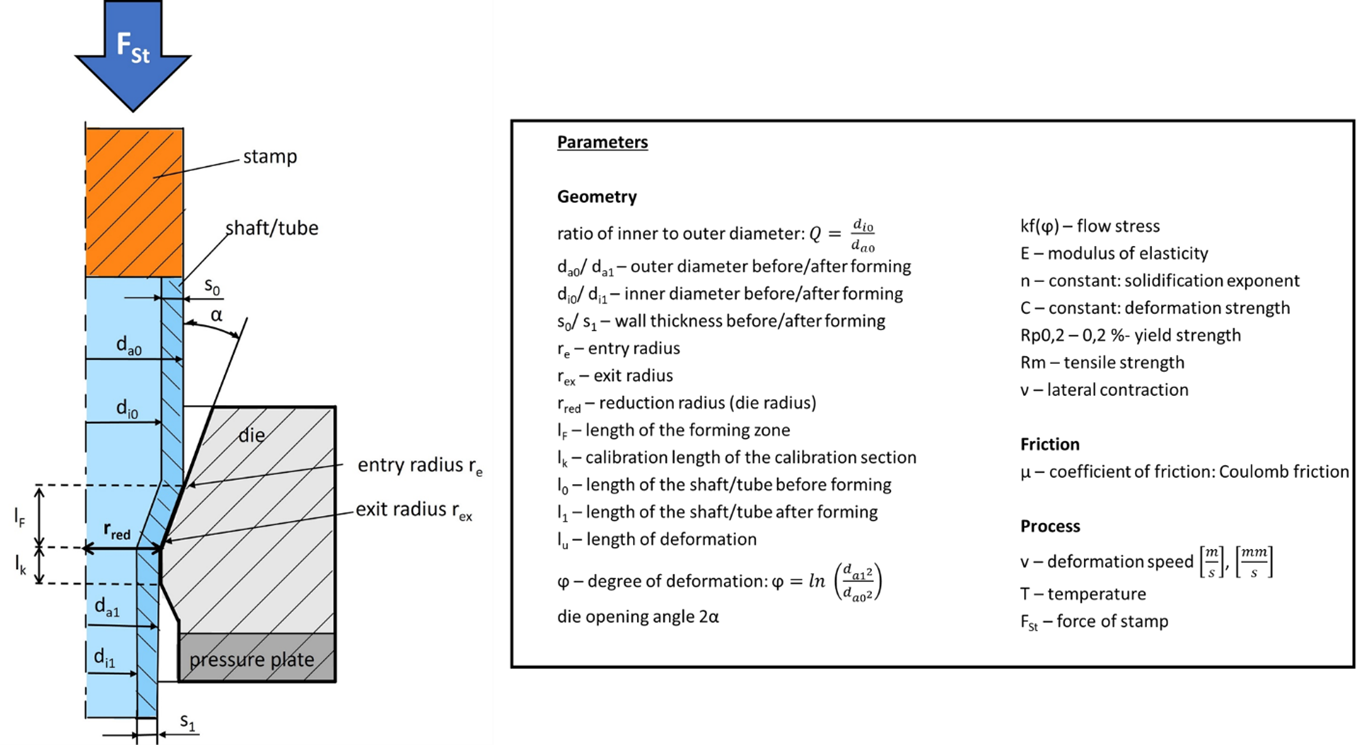}
\caption{Nosing of tubes without a Mandrel – principle and key parameters based on \cite{Handbuchumformen}.}
\label{fig:nosing_without_mandrel}
\end{figure}

The DIN 8583-6 defines the Manufacturing processes of forming by forcing through an orifice (extrusion), listing the nosing of tubes under the classification number 2.1.5.1 in the category of 'Extrusion'\cite{DIN8583-6}. 
In the nosing process, the stamp exerts a force ($F_{st}$) to move the tube towards the die, pressing it into the conical section of the die. During this process, the raw part is processed in the direction of the inlet cone, and its external diameter ($d_{a,0}$) is reduced to the exit diameter ($d_{a,1}$) as it exits the die, as illustrated in \autoref{fig:nosing_without_mandrel}. Both the inlet radius ($r_e$) and the outlet radius ($r_{ex}$) are formed naturally. These should not be confused with the reduction radius ($r_{red}$), which defines the reduction diameter within the die and is a parameter of the tool design, hence tool-related. This reduction diameter is equipped with a rounding ($Rm$) to facilitate the unhindered flow of the material – this rounding essentially serves as a radius and acts like a rounded bending edge \cite{Ebe80}.

During the forging process, a force is applied axially by  the a stamp to the tube. This force causes deformation not only in the contact zone between the workpiece and the die cone but also beyond this zone. The length of the forming zone $l_{F}$ pertains solely to the contact zone between the die surface and the workpiece. It is calculated using \cite{Alb90}:

\begin{align}
    l_{F} = \frac{r_{ex0} - r_{red}}{\tan(\alpha)}.
\end{align}

This length is defined as the forming zone and is related to the die. Another term used to describe the forging operations is the deformation zone, which refers to the workpiece. The deformation zone is ``the area of the workpiece that is in a plastic state due to the influence of the forming zone`` \cite{Alb90}. 

The wall thickness varies along the length of the component and forms freely. To deliberately control the change in wall thickness, experience and a sufficient number of trials for specific outer and inner diameters are required. However, even then, the alteration in wall thickness remains a free-forming process, with its consistent setting being only limitedly feasible \cite{Alb90}, \cite{Haa83}.

\subsubsection{Process limits}
\begin{figure}[h!]
\centering
\includegraphics[width=1\textwidth]{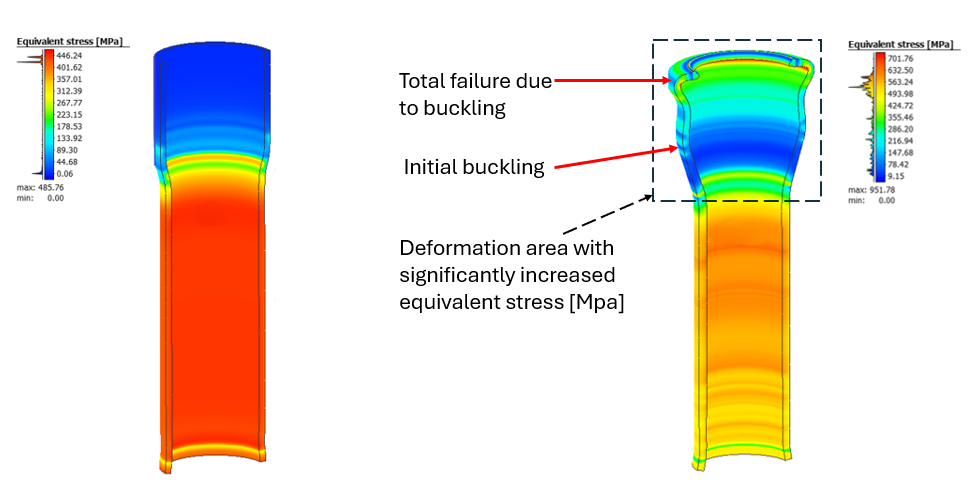}
\caption{Process limits. Picture left: Nosed tube with non-deformed shape (maximum equivalent stress: 485 MPa). Picture right: Nosed tube with cross folds caused by excessive axial compressive stresses (maximum equivalent stress: 951 MPa), which lead to initial buckling, total failure due to buckling, and a deformation area exhibiting significantly increased equivalent stress.}
\label{fig:Rohre}
\end{figure}

In the nosing process of tubes, the limits of the procedure are determined more by the material's capacity to deform rather than by the permissible load on the die. During the nosing of tubes, the forces exerted on the die are relatively low – it is more likely for the workpiece to fail before the die becomes overloaded. The common causes of failure include:
\begin{itemize}
    \item Buckling
    \item Collapsing/Buckling Outward (Cross folds - perpendicular to the axis of the workpiece) (see \autoref{fig:Rohre})
    \item Longitudinal folding – along the axis of the workpiece
    \item Cracks at the formed end of the tube  \cite{Haa83}, \cite{Ebe80}
\end{itemize}

Buckling typically occurs at and above the inlet radius, marked by an increase in the workpiece diameter, ranging from barely noticeable to pronounced bulging. The inlet radius is critical as it is where longitudinal and bending stresses overlap, exacerbating the curvature in the material. Failure through buckling is imminent when longitudinal stress alone reaches the material's flow stress, meeting  Tresca's flow criterion \cite{Lipp81}:

\begin{align}
    \sigma_{\text{max}} - \sigma_{\text{min}} = k_{f0}
\end{align}

Cross folds are caused by excessive axial, and longitudinal folds by high tangential compressive stresses. These folds predominantly occur in thin-walled workpieces with a diameter-to-wall-thickness ratio exceeding $d_{ao}/s_0 >60$, while thick-walled tubes with a ratio between $4 <  d_{a0}/s_0 <10$ are less prone to such folding \cite{Haa83}. Buckling in the inlet radius can lead to increased axial stress, resulting in cross folds. Particularly in thin-walled tubes, the likelihood of failure due to buckling under increasing compressive stress is higher compared to buckling failure.

Outward buckling occurs in the unformed area of the workpiece when the relative buckling force exceeds the allowable buckling stress, as shown in \autoref{fig:Rohre}. The permissible buckling stress depends on the slenderness and wall thickness of the workpiece, as well as the material's Young's modulus. For centrally applied forces, Euler's formula applies for the critical buckling stress and slenderness ratio.

\begin{equation}
\sigma_k = \frac{\pi^2 \cdot E}{\lambda^2} \label{eq:2.4}
\end{equation}
with
\[
\lambda = \frac{l_K}{\sqrt{\frac{I}{A_o}}}.
\]

In most nosing cases, the third Euler case of loading is assumed. For particularly slender tubes, buckling stress can be determined using the yield strength.
Cracks at the formed end of the sample are induced by tangential tensile stress resulting from the forging process, a portion of which remains as residual stress in the deformed workpiece. This cause of crack formation is attributed to tangential tensile stress.
Another significant factor determining the limits of the process is friction. Higher friction values increase the total force required for forging. Consequently, this reduces the achievable degree of forming and increases the likelihood of failure cases \cite{Alb90}.

\subsubsection{Dimensional accuracy}
\begin{figure}[h!]
\centering
\includegraphics[width=0.6\textwidth]{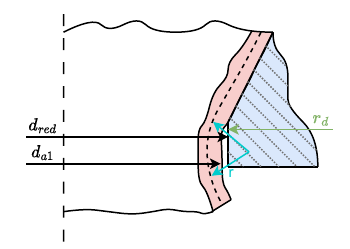}
\caption{Independent formation of a bending radius $r$ \cite{Haa83}. Tube coloured in red, die coloured in blue.}
\label{fig:bending_radius}
\end{figure}

In the nosing process without a mandrel, characteristic dimensional deviations occur, which can lead to significant tolerance exceedances. Once the material initially passes through the cone, it reaches the die radius area  (area of the rounding radius of the die \textit{r\textsubscript{d}}), intended to facilitate a smooth transition from the cone to the reduction diameter. At this point, the material detaches from the die and independently forms a bending radius \textit{r}, which also extends around the calibration section, effectively nullifying its calibration function, as illustrated in \autoref{fig:bending_radius}.

\begin{figure}[h!]
\centering
\includegraphics[width=1\textwidth]{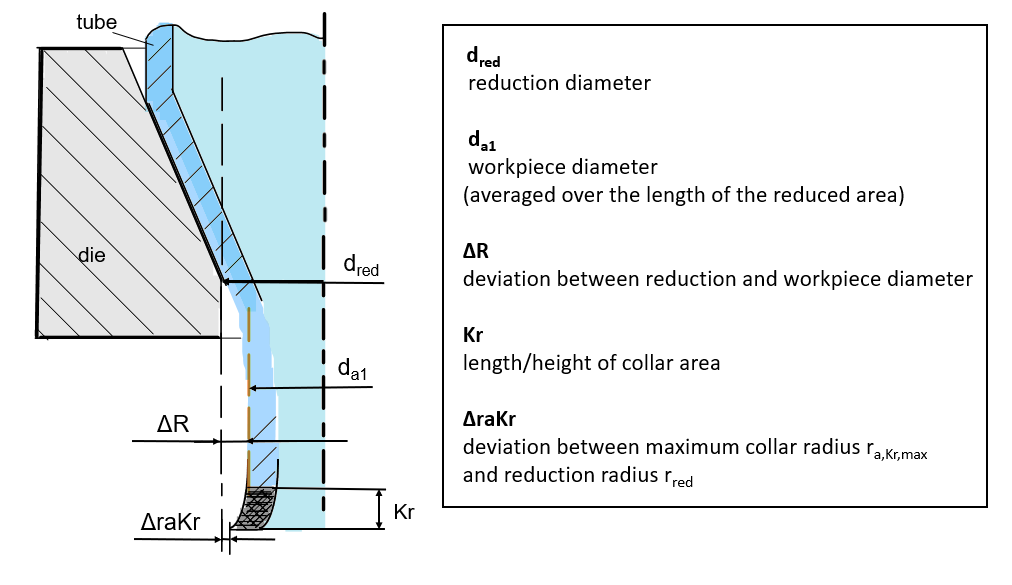}
\caption{Sizes of the dimensional deviation after forging  \cite{Alb90}.}
\label{fig:my_label3}
\end{figure}

It is observed that the cause of this detachment could be attributed to an inadequately small rounding radius\textit{ r\textsubscript{d}}. A larger rounding radius \textit{r\textsubscript{d}} might improve dimensional accuracy \cite{Haa83}. On the other hand, the conventional conical die contour is considered to create a stress state in the forming zone that causes the material to detach from the die \cite{Alb90}.
A result of the self-adjusting bending radius \textit{r} is the undershooting of the required target diameter, leading to greater deformation than initially planned in the process design. It is noted that achieving the target degree of forming with the initial process design is not feasible, requiring multiple trials and empirical data to determine a suitable reduction diameter \textit{d\textsubscript{red}} that can produce workpieces with the required external diameter \cite{Alb90}. The dimension of the deviation between the target reduction diameter \textit{ d\textsubscript{red}} and the actually formed external diameter \textit{ d\textsubscript{a1}} can be described as $\Delta R$ and is shown in \autoref{fig:my_label3}  \cite{Alb90}. Here\textit{ d\textsubscript{a1}} is determined as the averaged value of several points along the formed cylindrical shank length  without including the collar area \cite{Haa83}:

\begin{align}
    \Delta R = d_{a1}- d_{red}.
\end{align}

Another critical factor for dimensional deviations in the nosing process is the usually larger outer diameter of the collar area (Kr) compared to the subsequent cylindrical shaft. An additional metric of interest is the distance $\Delta ra_{\text{KR}}$, which represents the difference between the maximum outer radius of the collar area and the die radius. A smaller difference indicates a more pronounced funnel shape of the collar area. Although this area typically falls outside the required dimensions and is generally trimmed off, it could become an important secondary characteristic in future studies.

\subsubsection{Wall thickness change during nosing}
\begin{figure}[h!]
\begin{subfigure}{0.5\textwidth}
    \includegraphics[width=\textwidth]{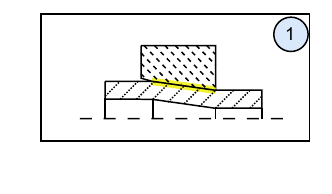}
    \caption{}
    \label{fig:wall_tickness_profile_literature_1}
\end{subfigure}
\begin{subfigure}{0.5\textwidth}
    \includegraphics[width=\textwidth]{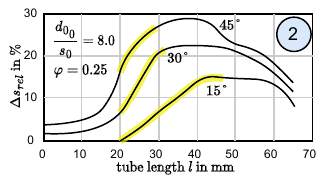}
    \caption{}
    \label{fig:wall_tickness_profile_literature_2}
\end{subfigure} \hfill
\begin{subfigure}{0.5\textwidth}
    \includegraphics[width=\textwidth]{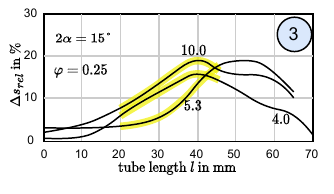}
    \caption{}
    \label{fig:wall_tickness_profile_literature_3}
\end{subfigure}
\begin{subfigure}{0.5\textwidth}
    \includegraphics[width=\textwidth]{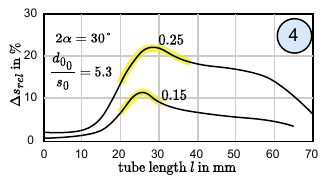}
    \caption{}
    \label{fig:wall_tickness_profile_literature_4}
\end{subfigure}
\caption{ Wall thickness profile in nosed tubes based on \cite{Haa83}. Yellow sections highlight contact between tube and die. (a): sketch of contact  section (b): The influence of the opening angle $2\alpha$ (c): The influence of the diameter-to-wall-thickness ratio $\frac{d_{ao}}{s_0}$ (d): The influence of the degree of forming $\varphi$.}
\label{fig:wall_tickness_profile_literature}
\end{figure}

\autoref{fig:wall_tickness_profile_literature} illustrates the wall thickness profiles, i.e., the relative change in wall thickness along the longitudinal axis of the sample. The figure schematically represents the longitudinal section of the workpiece and die at the end of the process. The yellow marking indicates the area in contact with the cone.

According to \cite{Haa83}, the wall thickness profile is influenced by various factors:
\begin{itemize}
    \item The influence of the opening angle $2\alpha$ results in an increase in the relative change in wall thickness $\Delta s_{\text{rel}}$.
    \item A larger diameter-to-wall-thickness ratio $\frac{d_{ao}}{s_0}$ also leads to an increase in $\Delta s_{\text{rel}}$.
    \item The degree of forming $\varphi$ also affects $\Delta s_{\text{rel}}$: A higher degree leads to an increase, while a lower degree results in a decrease in wall thickness change.
\end{itemize}

The profiles shown do not display constant sections of wall thickness change, an observation attributed to the short length of the samples. It is suggested that with longer samples, constant phases of the quasi-stationary forming zone could emerge. The impact of friction conditions on wall thickness change is also emphasized. Unfavorable friction conditions can lead to an increase in wall thickness and a decrease in length change, particularly in thin-walled workpieces \cite{Haa83}.

\subsection{Finite element models}
Now that the specific influences on the change in wall thickness during nosing have been investigated in detail, the focus is now on modelling these processes. The FEM-simulation of such processes is crucial in order to be able to predict the behaviour of the tubes under different conditions.
\subsubsection{Model description}
\begin{figure}[h!]
\centering
\includegraphics[width=1\textwidth]{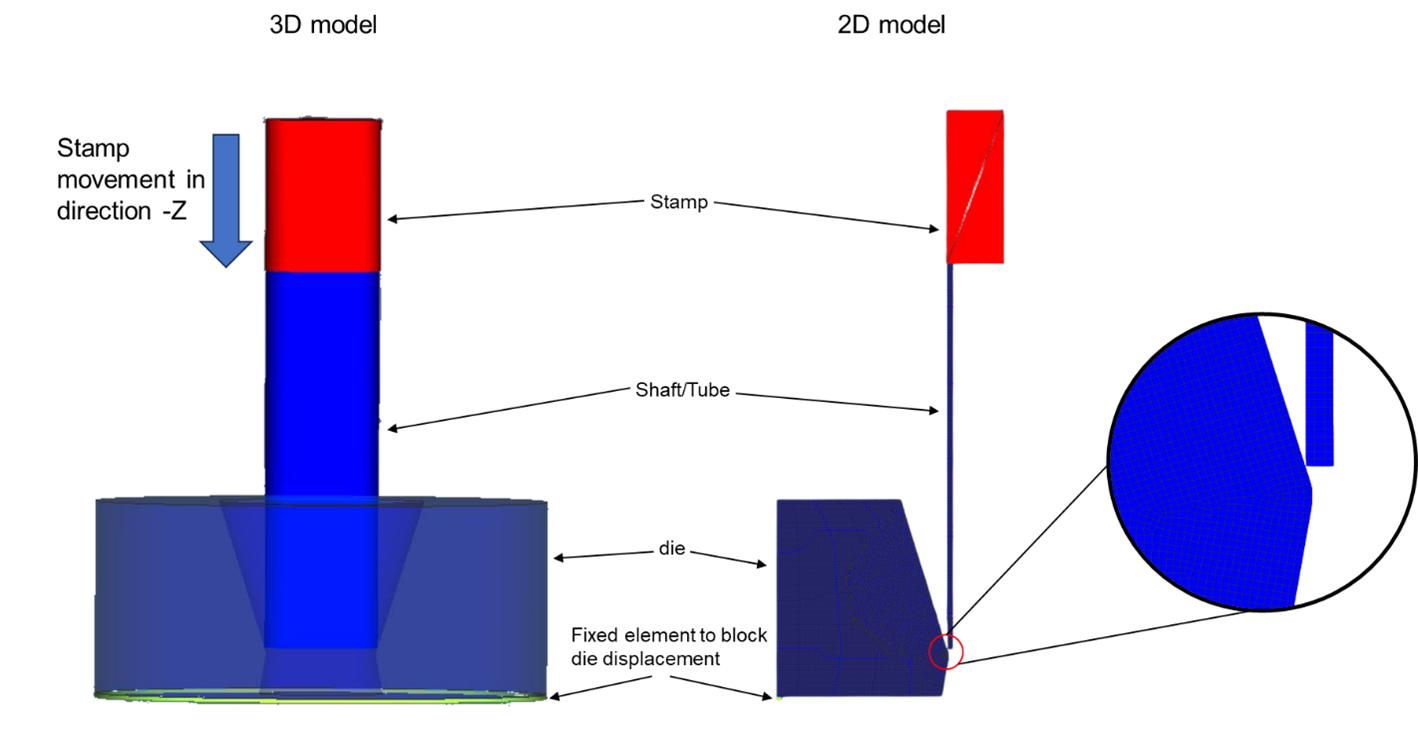}
\caption{Representation of the 3D and 2D finite element model.}
\label{fig:3d_2d_model}
\end{figure}

FEM simulations are used to generate training data for the neural network model. This data represents the change in wall thickness under different parameter changes.
In this study, the simulation process was modeled with Simufact Forming as a cold forging operation using two tools. Calculations were carried out on a 2D axisymmetric plane instead of a 3D volume body due to the radial symmetry of all parts. The stamp was defined as a rigid element, while the tube and the die were designated as deformable bodies. Additionally, another rigid element was positioned below the die to ensure its stabilization (see \autoref{fig:3d_2d_model}). Since this was only used to fix the die in the \ac{FEM} simulation, this rigit element is not part of the data export and will not be visibible in any of the later results. 

In the FEM model, the material properties for the alloy were sourced from a material database from simufact. The material is a Copper-Zinc-Lead alloy, known as CuZn39Pb2 or CW612N. This alloy is characterized by a valid temperature range on the material card from 20°C to 300°C. The Young's modulus of the alloy is noted to be 105,000 MPa. The flow stress, denoted as $kf$, is calculated using Hollomon's-formula 
\begin{equation}
kf = \sigma_F = C \cdot \varphi^n,
\end{equation}
where the strain hardening exponent $n$ is 0.334, and the constant $C$, representing the material's strength coefficient, is 794.965 MPa. This coefficient, along with the exponent, is crucial in determining the tensile strength ($R_m$), which is approximately 395 MPa.

For simulating the forging process, a hydraulic press was chosen to define the kinematics, facilitating a linear stamp movement in the negative $Z$-direction at a constant velocity of 1000 mm/s. The FEM model incorporates the Coulomb friction model with a constant friction coefficient $\mu$ to idealize the friction factor. In the simulations carried out, the friction coefficients used were 0.05 and 0.2. The tool's temperature settings are defined with an initial constant temperature of 20\textdegree C and a heat transfer coefficient to the environment of 50 W/(m\textsuperscript{2}·K). Simultaneously, the workpiece's temperature is set at a constant 20\textdegree C with an identical heat transfer coefficient to the environment.

Meshing the workpiece body involved selecting the 'Advancing Front Quad' technique for creating 2D geometries with quadrilateral elements. This meshing method predetermined the use of Quad (10) elements, suitable for axisymmetric bodies.

\subsubsection{Mesh convergence study}
The increase in wall thickness can be represented both in absolute and relative terms in order to quantitatively describe the change in material thickness in a forging process.
The absolute wall thickness increase $\Delta s$ is expressed as the difference between the final thickness after the forging process $s_1$ and the initial thickness $s_0$ of the material. It is specified in the same unit as the wall thickness, in millimeters. It is calculated as follows:
\begin{equation}
\Delta s = s_1 - s_0.
\end{equation}

The relative increase in wall thickness $\Delta s_{\text{rel}}$ is given as a percentage and expresses the change in wall thickness in relation to the original thickness. It is calculated as follows:
\begin{equation}
\Delta s_{\text{rel}} = \frac{s_1 - s_0}{s_0} \cdot 100\% = \frac{\Delta s}{s_0} \cdot 100\%,
\end{equation}
with:
\begin{itemize}
    \item $\Delta s_{\text{rel}}$ - relative wall thickness change [\%]
    \item $s_0$ - wall thickness of the tube before forging [mm]
    \item $s_1$ - wall thickness of the tube after forging [mm]
\end{itemize}

\begin{figure}[h!]
\begin{subfigure}[b]{0.5\textwidth}
    \includegraphics[width=1\textwidth]{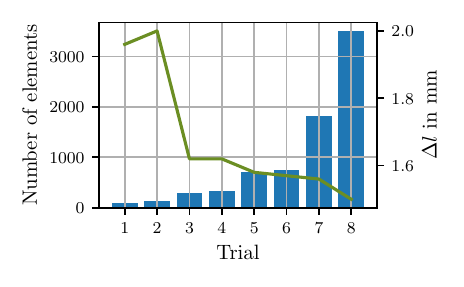}
    \caption{}
\end{subfigure}
\begin{subfigure}[b]{0.5\textwidth}
    \includegraphics[width=1\textwidth]{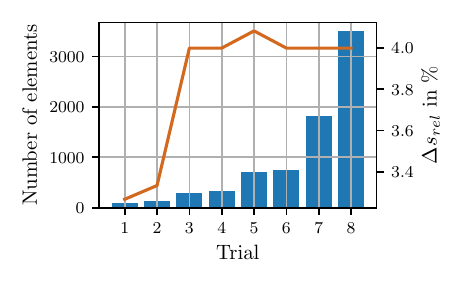}
    \caption{}
\end{subfigure}
\caption{Results of mesh convergence study.}
\label{fig:konvergenzstudie}
\end{figure}

The relative wall thickness change in percent serves as a standardized value that makes it possible to compare the wall thickness increase independently of the original wall thickness $s_0$ of the material before forging. This normalized value is particularly useful because it puts the change in wall thickness in relation to the original thickness and thus provides a comparable value, regardless of the absolute dimensions of the starting material. By standardizing to the original wall thickness $s_0$, the relative change in wall thickness $\Delta s_{rel}$ is converted into a percentage form that is easy to interpret and compare.
For the mesh convergence study, the relative change in wall thickness in percent $\Delta s_{\text{rel}}$ and the absolute change in length in millimeters $\Delta l$ were considered (see \autoref{fig:konvergenzstudie}). The relative change in wall thickness $\Delta s_{\text{rel}}$ was the decisive variable, and the absolute change in length $\Delta l$ served as a further comparative variable.

The element size varied from 1.0 millimeters to 0.2 millimeters. From an element size of 0.4 millimeters, there were no significant changes in the relative change in wall thickness $\Delta s_{\text{rel}}$, which indicates that mesh convergence had been achieved. This element size was therefore selected for the simulations.

Although a finer mesh could provide even more accurate results for the absolute change in length $\Delta l$, this would involve significantly longer computing time. The choice of an element size of 0.4 mm therefore represents an optimal balance between accuracy and computational effort as shown in \autoref{tab:mesh_convergence}.

\begin{table}[h!]
    \centering
    \caption{Table of results of mesh convergence study.}
    \label{tab:mesh_convergence}
    \begin{tabular}{c | d{3.1} | d{4.0} | d{1.2} | d{1.2}}
    \toprule
    Trial & \mc{Element Size} & \mc{Number of Elements} & \mc{$\Delta s_{rel}$ in \%} & \mc{$\Delta l$ in mm} \\
    \midrule
    1 & 1.0 &  100 & 3.27 & 1.96 \\
    2 & 0.8 &  125 & 3.33 & 2.00 \\
    3 & 0.7 &  284 & 4.00 & 1.62 \\
    4 & 0.6 &  332 & 4.00 & 1.62 \\
    5 & 0.5 &  699 & 4.08 & 1.58 \\
    6 & 0.4 &  750 & 4.00 & 1.57 \\
    7 & 0.3 & 1812 & 4.00 & 1.56 \\
    8 & 0.2 & 3500 & 4.00 & 1.5 \\
    \bottomrule
    \end{tabular}
\end{table}

\subsubsection{Parameters}
Based on previous analysis, three key parameters will be investigated to determine their impact on the dimensional deviation $\Delta R$ and wall thickness change:
\begin{enumerate}
    \item the degree of deformation ($\varphi$),
    \item the die opening angle ($\alpha$),
    \item and the Coulomb friction coefficient ($\mu$).
\end{enumerate}

While other variables remained constant, these three parameters were varied to assess their specific effects on the nosing process. The length of the calibration stretch and the diameter ratio $Q$ were maintained constant, focusing the analysis solely on the impacts of $\varphi$, $\alpha$, and $\mu$ as shown in \autoref{tab:simulation_run_parameters}. Using this specific split of training and testing data, the total number of timesteps ends up in 207778 for our training data and 48045 for our testing data.

\begin{table}[h!]
    \caption{Simulated runs based on the parameters the degree of deformation $\varphi$, the opening angle $\alpha$, and the coulomb friction coefficient $\mu$. Blue cells are used for training and red cells are held out for testing. FEM simulations from the gray cells resulted in deformed workpieces out of specification. Gray and green cells were not used in training or testing.}
    \label{tab:simulation_run_parameters}
    \centering
    \begin{tabular}{c|c|C{0.5cm}|C{0.5cm}|C{0.5cm}|C{0.5cm}|C{0.5cm}|C{0.5cm}|C{0.5cm}|C{0.5cm}|C{0.5cm}}
    \multicolumn{2}{c}{} & \multicolumn{9}{|c}{$\alpha$} \\
    \toprule
    $\mu$ & $\varphi$ & 5° & 7.5° & 10° & 12.5° & 15° & 17.5° & 20° & 22.5° & 25° \\
    \midrule
    \multirow{7}{*}{0.05} & 0.05 & \cellcolor{colortrain} & \cellcolor{colortrain} & \cellcolor{colortrain} & \cellcolor{colortrain} & \cellcolor{colortrain} & \cellcolor{colortrain} & \cellcolor{colortrain} & \cellcolor{colortrain} & \cellcolor{colortrain} \\
   & 0.10 & \cellcolor{colortrain} & \cellcolor{colortrain} & \cellcolor{colortrain} & \cellcolor{colortrain} & \cellcolor{colortrain} & \cellcolor{colortest} & \cellcolor{colortrain} & \cellcolor{colortest} & \cellcolor{colortrain} \\
    & 0.15 & \cellcolor{colortrain} & \cellcolor{colortrain} & \cellcolor{colortest} & \cellcolor{colortrain} & \cellcolor{colortest} & \cellcolor{colortrain} & \cellcolor{colortest} & \cellcolor{colortrain} & \cellcolor{colortrain} \\
    & 0.20 & \cellcolor{colortest} & \cellcolor{colortrain} & \cellcolor{colortrain} & \cellcolor{colortest} & \cellcolor{colortrain} & \cellcolor{colortrain} & \cellcolor{colortrain} & \cellcolor{colortrain} & \cellcolor{colortrain} \\
    & 0.25 & \cellcolor{colortrain} & \cellcolor{colortrain} & \cellcolor{colortest} & \cellcolor{colortrain} & \cellcolor{colortrain} & \cellcolor{colortrain} & \cellcolor{colortrain} & \cellcolor{colornotperformed} & \cellcolor{colornotperformed} \\
    & 0.30 & \cellcolor{colortrain} & \cellcolor{colortrain} & \cellcolor{colortrain} & \cellcolor{colortrain} & \cellcolor{colornotperformed} & \cellcolor{colornotperformed} & \cellcolor{colornotperformed} & \cellcolor{colornotperformed} & \cellcolor{colornotperformed} \\
    & 0.35 & \cellcolor{colornotperformed} & \cellcolor{colortest} & \cellcolor{colortrain} & \cellcolor{colornotperformed} & \cellcolor{colornotperformed} & \cellcolor{colornotperformed} & \cellcolor{colornotperformed} & \cellcolor{colornotperformed} & \cellcolor{colornotperformed} \\
    \midrule
    \multirow{4}{*}{0.20} & 0.05 & \cellcolor{colornottrained} & \cellcolor{colornottrained} & \cellcolor{colornottrained} & \cellcolor{colornottrained} & \cellcolor{colornottrained} & \cellcolor{colornottrained} & \cellcolor{colornottrained} & \cellcolor{colornottrained} & \cellcolor{colornottrained} \\
    & 0.10 & \cellcolor{colornottrained} & \cellcolor{colornottrained} & \cellcolor{colornottrained} & \cellcolor{colornottrained} & \cellcolor{colornottrained} & \cellcolor{colornottrained} & \cellcolor{colornottrained} & \cellcolor{colornotperformed} & \cellcolor{colornotperformed} \\
    & 0.15 & \cellcolor{colornottrained} & \cellcolor{colornottrained} & \cellcolor{colornottrained} & \cellcolor{colornottrained} & \cellcolor{colornottrained} & \cellcolor{colornotperformed} & \cellcolor{colornotperformed} & \cellcolor{colornotperformed} & \cellcolor{colornotperformed} \\
    & 0.20 & \cellcolor{colornotperformed} & \cellcolor{colornottrained} & \cellcolor{colornottrained} & \cellcolor{colornottrained} & \cellcolor{colornottrained} & \cellcolor{colornotperformed} & \cellcolor{colornotperformed} & \cellcolor{colornotperformed} & \cellcolor{colornotperformed} \\
    \bottomrule
    \end{tabular}
\end{table}

For the training phase, parameter sets indicated by blue markers were utilized, whereas red indicates parameter sets used as the test datasets. The gray marked entries demonstrated issues such as upsetting (see \autoref{fig:Rohre}). Gray and green marked simulations were not employed in the training of the neural network model.

\section{Graph Neural Network Architecture to approximate FEM Simulations}
\label{sec:architecture}

In this section, we present a novel graph neural network architecture designed to approximate FEM simulations while reducing computation times considerably to allow for real-time applications of the model. 

\subsection{Network Requirements and challenges}

We analyse the specific requirements to the neural network architecture given the considered nosing process which yield to the novel graph neural network architecture presented in the subsequent section.

First, a particular property of the nosing process stems from the different interaction mechanisms. More specifically, the nosing process consists of three different types of physical contact relations. Particularly, we have to model contacts between moving objects (the stamp and tube), contacts between a moving and a static object (the die and tube) and the deformations within the tube itself. This yields to three different types of dynamics which we have to account for in the architecture design.

Second, we have to account for the dynamic meshing due to the tube movement within the process. More specifically, due to the tube's movement, different edges between the tube and the stamp mesh have to exchange information to be able to learn the physical relations. We solve this by dynamically adding edges to the grid based on the distance between nodes.

Third, to describe the accuracy of the learned network behaviour, we found standard measures like the mean square error as not fully convincing. Hence, we propose the \ac{ABTC}, which allows to better evaluate the model quality.

Subsequently, we will present our approach with the specific solutions to the above challenges.

\subsection{Graph Neural Network Architecture}

\begin{figure}[h!]
\centering
\includegraphics[width=1\textwidth]{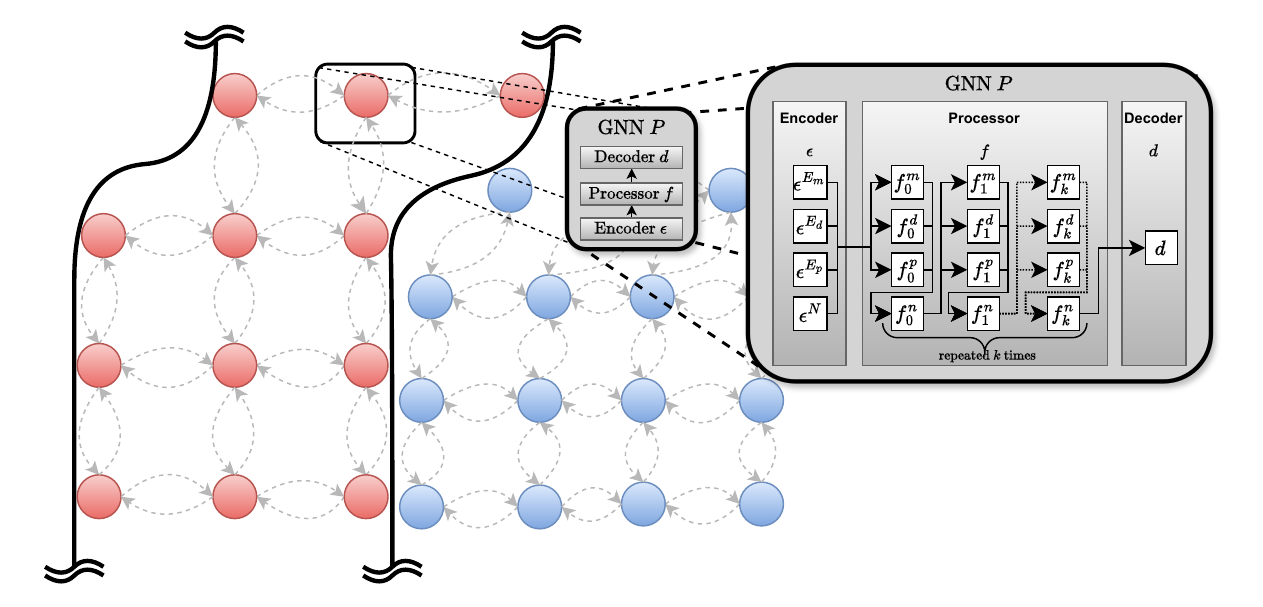}
\caption{Overview of how the GNN is linked to the nodes and edges in the mesh. Red nodes represent tube nodes, blue nodes represent nodes of the die. Each arrow represents one unidirectional edge.}
\label{fig:overview_gnn}
\end{figure}

There are many approaches and architectures for \acp{GNN}, as discussed in e.g. \cite{Waikhom2023, Khemani2024}, whereby we follow the specific approach from \cite{Pfaff2021}.
The architecture of the novel GNN including how the individual networks and sub-networks are linked to edges and nodes is shown in \autoref{fig:overview_gnn}. For both the tube as well as the die, most nodes are modeled with 4 neighbors, except for nodes on the boundaries of the mesh which have three or two. It is important to note, that each object (tube, stamp, die) is a closed mesh and does not contain edges to different objects. 
 
To be able to predict the deformation of the current graph, each node contains the information whether it is a solid node (the die and the stamp), or a deformable node (the tube). Additionally, we include the information of graph connectivity and the current positions of each node. To model deformations with different friction coefficients, we store the friction coefficient $m$ as an edge feature. 

We then parameterize a neural network $P$ that predicts the change of node positions given the local information of a node and all its connected neighbors. In order to predict the change in node positions, we follow the graph based encode-process-decode approach model of \cite{Pfaff2021}. Specifically, we construct a mesh $M=(N, E^m)$ composed out of nodes $N$ and of edges $E^m$. The objective of the \ac{NN} $P$ is to predict the change of $M^t \rightarrow M^{t+1}$ ($M^{t+1} = P(M^t)$). The time trajectory is then generated using an integrative approach starting with an initial mesh $M^0$. Hence, the model itself can be seen as a one-step predictor. The exact same model $P(M)$ is used for every node in the entire graph, i.e. the parameters are shared among the network nodes and edges, see \autoref{fig:overview_gnn}.

As mentioned before, the nosing process consists of contacts between two moving objects, moving and a static object (die and tube), and deformations in the tube itself. This results in three different edge types and considerably different edge dynamics. Particularly, we have to consider fundamentally different mechanisms and dynamics: one is an axial stress which causes the feed while the other is a radial and tangential stress causing the diameter and thickness change during the forging process.

Hence, using a single network for all edge types is not appropriate. Consequently we propose to define different encoders for the different interaction types. Particularly, we define a neural network for the tube edges, one for the dynamically added edges between the tube and the die, and one for the added edges between the stamp and the tube, as shown in \autoref{fig:overview_gnn}). Using a message passing algorithm with $k$ steps allows nodes to transmit information to, or receive information from, all nodes that can be reached within $k$-edges, since the graph-embedded features get updated at every step of the processor. We parameterize the edges $E$ with the relative cartesian coordinates as well as the euclidean distance of the connected nodes.

This results in four encoders $\epsilon$ in total, $\epsilon^{N}$ for the nodes and $\epsilon^{E_{d}}, \epsilon^{E_{t}}, \epsilon^{E_{p}}$ for the respective edge sets. The encoders are MLPs whose outputs are calculated following
\begin{equation}
    \epsilon^\square = 
    \text{LayerNorm} \left(
    \mathbf{W}_n
    \max(\mathbf{W}_{n - 1}( 
        \cdots 
            \max(\mathbf{W}_1\mathbf{x} + \mathbf{b}_1,  0) 
            + \cdots
        ) 
        + \mathbf{b}_{n - 1},
    0)
    + \mathbf{b}_n \right)
\end{equation}
whereby x denotes the respective encoder input.

Consequently, the processor \cite{Pfaff2021} is modified for three edge sets such that
\begin{equation}
    {e^\prime}_{i, j}^{d} \leftarrow f^d \left( e_{i, j}^{d}, n_i, n_j \right),~~
    {e^\prime}_{i, j}^{t} \leftarrow f^t \left( e_{i, j}^{t}, n_i, n_j \right),~~
    {e^\prime}_{i, j}^{p} \leftarrow f^p \left( e_{i, j}^{p}, n_i, n_j \right),~~
\end{equation}
with $e_{i, j}^{d} \in E_{d}$, $e_{i, j}^{t} \in E_{t}$, $e_{i, j}^{p} \in E_{p}$ and $n_i \in N$.
The node attributes are updated based on
\begin{equation}
    n_i^{\prime} \leftarrow f^n \left( 
        n_i, \sum_j {e^\prime}_{i, j}^{d}, \sum_j {e^\prime}_{i, j}^{t}, \sum_j {e^\prime}_{i, j}^{p} 
    \right) .
\end{equation}
Within the processor, $f^d$, $f^d$, $f^p$ and $f^n$ are also MLPs, connected with an additional residual connection.

A final decoder MLP $d$ decodes the node features $N$ produced by the processor to the desired model output.
As the target, we chose the difference $\Delta \mathbf{p} = \mathbf{p}^{t+1} - \mathbf{p}^{t}$ whereby $\mathbf{p}=(x, y)$ consists of the node coordinates. These targets are normalized as in \cite{Pfaff2021}.

\begin{figure}[h!]
    \centering
    \includegraphics{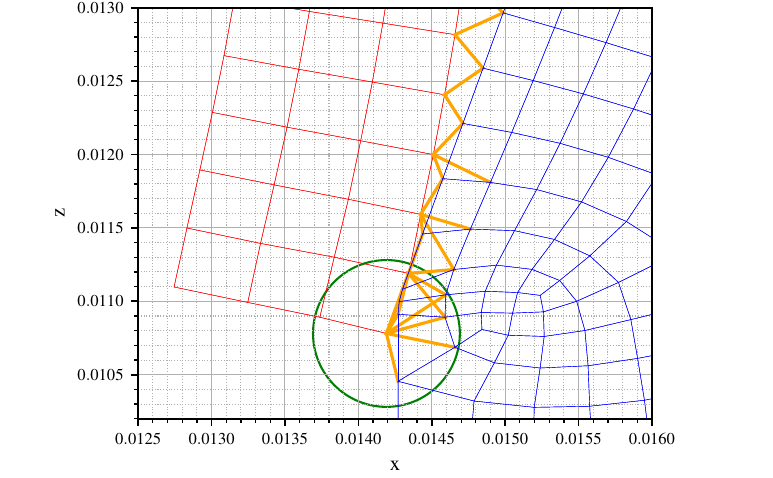}
    \caption{Visualization of the dynamically added edges for r=0.5 mm. The tube is shown in red, the die is shown in blue and the added edges in orange. The circle further illustrates the used radius $r$ for one node.}
    \label{fig:contact_transforms_visu}
\end{figure}

As stated before, we have to account for the dynamically moving objects resulting in varying interactions among the different subnetworks. To this end, for each step $t^n \rightarrow t^{n+1}$, we further dynamically add edges $E_{d}$ between nodes of the tube-mesh and the die-mesh as well as edges $E_{p}$ between the tube-mesh and the stamp-mesh, whose euclidean distance is smaller than $r$ whereby $r$ is a hyperparameter, as depicted in \autoref{fig:contact_transforms_visu}.

This enables the message-passing process between the otherwise unconnected sub-graphs of the tube and the die as well as the tube and the stamp. 

We note, that the adaptive re-meshing \cite{Pfaff2021} is not required for our application as the mesh formed by the FEM system already contains the necessary node resolution. Since we performed this convergence analysis in the FEM simulations, we waive on using adaptive re-meshing. 

\subsection{Performance Metrics}

To evaluate the performance of the predicted sequences, we have to define corresponding metrics. We start with the common \ac{RMSE}, which however, lacks interpretability in our application scenario. Hence, we subsequently design a novel metric with better interpretability.

\subsubsection{Root Mean Square Error}
As our first metric, we calculate the \ac{RMSE} for all nodes of the tube. It is calculated as 
\begin{equation}
    RMSE = \sqrt{\frac{1}{N}\sum_{i=1}^{N} \lVert y_i - x_i \rVert^2}
\end{equation}
where $y_i$ is the predicted node coordinate and $x_i$ the FEM node coordinate. This metric is useful as it gives information about the distance between the respective ground truth and predicted node coordinates and lets us evaluate the susceptibility to error propagation and potential exponential increase over the rollout steps. However, it lacks insights about the result of the forging process and is hardly interpretable. Thus, we additionally introduce a second metric.

\subsubsection{Area between the Thickness Curves}

Even though an analysis of the wall thickness change in cold forging processes is common practice \cite{ALVES2009521}\cite{ALASWAD2011838}\cite{Alves2009}, there does not seem to be a common metric comparing the accuracy of models. Hence, we introduce the \ac{ABTC} as a second evaluation metric. As we are mainly interested in the thickness / thickness change of the tube, this gives us a valuable further metric to rate the performance of the predicted results. 

For a specific timestep $t$, consider $f(l)$ and $g(l)$ as the functions defining the tube thickness over its length $l$ for the \ac{NN} as well as the FEM results, divided into $k$ segments by their intersection points with their respective start and end points $\{(a_0, b_0), (a_1, b_1), \dots, (a_k, b_k)\}$ whereby $a_0 = 0$, $a_i = b_{i - 1}$ and $b_k = l$. Then, the \ac{ABTC} is calculated  as
\begin{equation}
    ABTC = \sum_{i=0}^{k} \lVert
        \int_{a_i}^{b_i} f(l) \,dl - \int_{a_i}^{b_i} g(l) \,dl
    \rVert .
\end{equation}
A lower area corresponds to a better prediction. We note, that we use the \ac{ABTC} for evaluation only.

\subsection{Implementation Details}

\subsubsection{Neural Networks Details}

The input mesh $M$ is first fed into the node and the three edge encoder respectively, whereby all encoder $\epsilon$ consist of three Linear-ReLU combinations with hidden size 128 and a final Linear layer with output size of 128 followed by a LayerNorm. 

The processor is build with $k=15$ sequential message passing steps, build based on the generalized GraphNet blocks from \cite{Pfaff2021} and connected with a residual connection. The MLPs $f^d$, $f^t$, $f^p$ and $f^n$ within these blocks share the architecture with the node and edge encoder $\epsilon^{N}$, $\epsilon^{E_{d}}, \epsilon^{E_{t}}~\text{and}~\epsilon^{E_{p}}$.

The processors output is then finally fed to the output decoder $d$, which has again the same architecture than the input encoders $\epsilon$, with the difference that the output size is two (one for each component of $\Delta p$) and no LayerNorm is used.

\subsubsection{Dataset Modifications}

\begin{figure}[htb]
    \centering
    \includegraphics{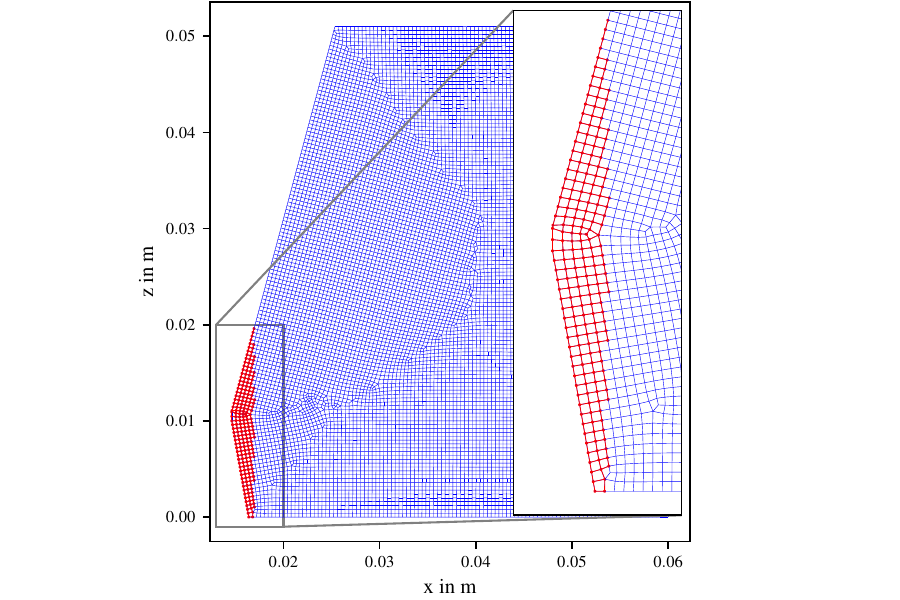}
    \caption{Visualization of the die. Blue shows the initial mesh as produced by the FEM simulation, red shows the remaining nodes after filtering to be used in the \ac{NN} training.}
    \label{fig:tool_reduction}
\end{figure}

The die obtained from the FEM simulation contains a huge amount of nodes due to its massive construction. As we model it as non-deformable within the \ac{NN} simulation, this is not required and instead causes a massive computational overhead. Thus, we reduce the nodes as indicated by \autoref{fig:tool_reduction}.

Specifically, we cut off nodes with $x > 0.017$ since we know the tube has an outer limit of $x = 0.015$. 

\subsubsection{Training Setup}

We train our model for 50 epochs or 370750 steps respectively using a batch-size of 28 and the MSE Loss function is applied to the node coordinates. As the die and stamp are non-deformable, we mask the output prediction such that only the tube-nodes get utilized for the loss calculation. The optimizer used is the Adam optimizer and the learning-rate is set to $1 \times 10^{-4}$ with an exponential decay to $1 \times 10^{-5}$. The contact radius $r$ is chosen as 0.8 mm.

In order to lower the susceptibility of the one-step predictor for error propagation issues during a multi-step rollout and increase prediction stability, we add noise during training, similar to \cite{pmlr-v119-sanchez-gonzalez20a, Pfaff2021}. Specifically, we determine the mean displacement in x- and z-direction of all tube nodes, multiply with a factor $1 \times 10^{-3}$ and use that as the noise standard deviation.
This noise is then added to the node coordinates and subtracted from the target vector, such that the desired node position at $t + 1$ is unaffected by it. We separate between x- and z noise since the mean displacement distance is $\approx-2.13 \times 10^{-7}\,\si{m}$ in x-direction and $\approx-1.50 \times 10^{-5}\,\si{m}$ in z-direction, thus differentiates by two magnitudes.

\section{Results}
\label{sec:results}

In the following, we present the obtained results using the FEM simulation as well as the \ac{NN} approach to approximate the FEM results.

\subsection{Results of finite element simulations}
The results of the FEM simulations illustrate the influence of the parameters on the change in wall thickness.

\begin{figure}[h!]
\centering
\includegraphics[width=0.6\textwidth]{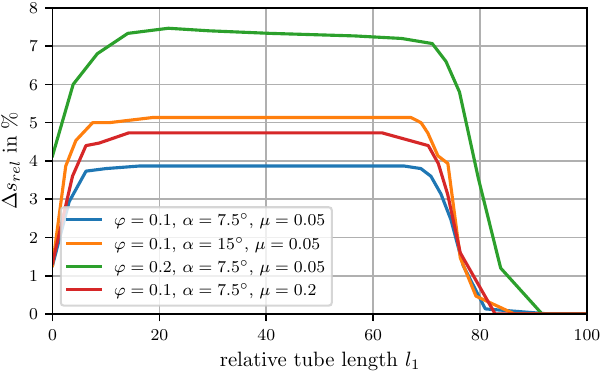}
\caption{Relative wall thickness change $\Delta s_{rel}$ depending on the degree of deformation $\varphi$, the die opening angle $\alpha$, and the friction coefficient $\mu$.}
\label{fig:wandaenderung}
\end{figure}

\autoref{fig:wandaenderung} shows the wall thickness profile when the respective parameters are varied. Doubling the degree of deformation here results in an relative increase in the wall thickness change from 3.7\% to 7.2\%, indicating an overall increase of approximately +93\%. Furthermore, it is observed that as the degree of deformation increases, the wall thickness alteration exhibits less constancy over its progression. The area of the quasi-stationary phase is notably shorter and demonstrates a slight inclination, indicating the absence of consistent strength properties. Additionally, at higher degrees of deformation, the initial slope is twice as long. Doubling the die opening angle here results in an increase from 3.7\% to 5.0\%, representing an overall rise of approximately +34\%. Quadrupling the coulomb friction coefficient here leads to an increase from 3.7\% to 4.6\%, signifying an overall augmentation of approximately +25\%.

\begin{figure}[h!]
\begin{subfigure}{0.5\textwidth}
    \includegraphics[width=\textwidth]{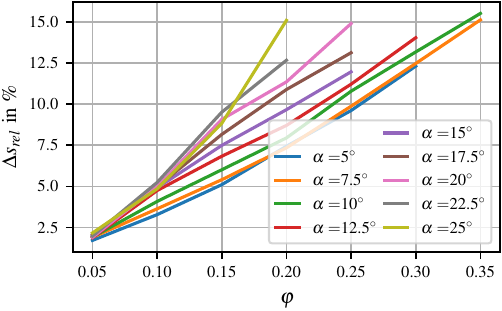}
    \caption{}
    \label{fig:relative_wall_thickness}
\end{subfigure}
\begin{subfigure}{0.5\textwidth}
    \includegraphics[width=\textwidth]{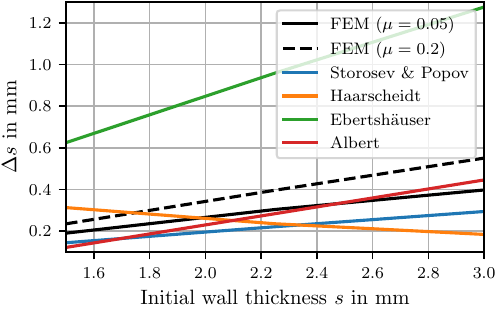}
    \caption{}
    \label{fig:comparison_approaches}
\end{subfigure}
\caption{(a):{Non-linearity of the wall-thickness increase} (b): Comparison of the predictive analytical models with FEM simulations }
\end{figure}

In \autoref{fig:relative_wall_thickness}, the relative wall thickness change $\Delta s_{\text{rel}}$ is plotted against the degree of deformation $\varphi$ for various die opening angles.When trying to show a possible linear relationship between the variables, it was noticed that the curves represent non-linear functions and therefore linearity is not given. For the size of the degree of deformation $\varphi$ plotted in \autoref{fig:relative_wall_thickness}, the diameter $d_{a1}$ calculated in the simulation software was used as the averaged value to calculate the degree of deformation using the following formula:
\begin{equation}
\varphi = \ln\left(\frac{d_{a0}}{d_{a1}}\right)^2 = 2 \ln\left(\frac{d_{a0}}{d_{a1}}\right).
\end{equation}

\begin{figure}[h!]
\centering
\includegraphics[width=0.6\textwidth]{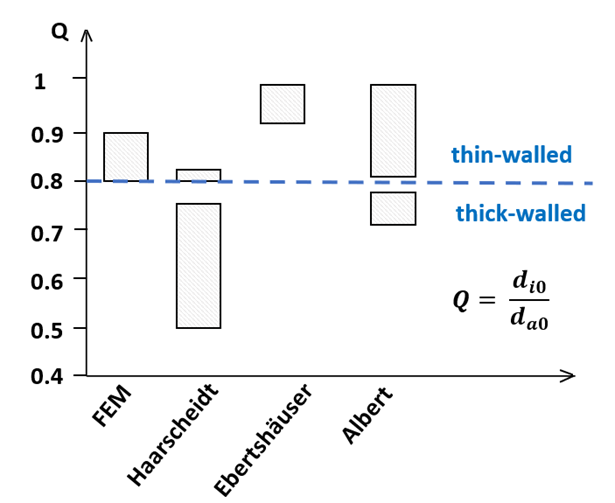}
\caption{Scope of validity of the models from the literature}
\label{fig:dick_dunnwand}
\end{figure}

\autoref{fig:comparison_approaches} compares the prediction models from the literature with FEM investigations with an outer diameter $d_{a0}$ of 30 mm and different initial wall thicknesses $s_0$ of 1.5 mm, 2.25 mm, and 3 mm were calculated for the comparison. The degree of deformation was $\varphi = 0.3$ and the Coulomb coefficient of friction was varied with $\mu = 0.05$ and $\mu = 0.2$.
The obtained observation shows that the models are inhomogeneous – both the predicted wall thickness increases $\Delta s$ differ greatly in height and orientation, but also the scope of validity of the models is divergent. The models from the literature each apply to a specific scope of validity for tubes with higher or lower wall thicknesses. In the literature, thin-walled and thick-walled tubes are classified with the size $Q$ – the diameter ratio of the inner diameter to the outer diameter of a tube – by arbitrarily determining that tubes with $Q > 0.8$ are considered thin-walled and tubes with $Q < 0.8$ are considered thick-walled, see \autoref{fig:dick_dunnwand}. The diameter ratio $Q$ is calculated as follows:

\begin{equation}
Q = \frac{d_{i0}}{d_{a0}}.
\end{equation}

Furthermore, none of the models from the literature take the influence of friction into account in their formulas, whereas the FEM model takes the influence of friction into account. Haarscheidt's empirical formula for predicting the change in wall thickness is as follows \cite{Haa83}, whereby he mainly examined thick-walled tubes in his tests:

\begin{equation}
\Delta s = \left(\sqrt{\frac{d_{a0}}{d_{a1}} - 1}\right) \cdot \left[\ln\left(\frac{d_{a0}}{s_0} / 0.6\right)\right]^{2.5} \cdot 0.2 \cdot \sqrt{\sin 2\alpha}.
\end{equation}

Ebertshäuser's analytical formula \cite{Ebe80} is as follows, whereby he only examined thin-walled tubes in his tests:

\begin{equation}
\Delta s = \frac{62}{\left(\frac{d_{a0}}{s_0}\right)} \cdot (\beta_e - 1).
\end{equation}

Instead of the degree of deformation $\varphi$, Ebertshäuser uses a so-called retraction ratio $\beta_e$ between the initial diameter before and after deformation \cite{Ebe80}:

\begin{equation}
\beta_e = \frac{d_{a0}}{d_{a1}}.
\end{equation}

In Albert's tests, various thin-walled and thick-walled tubes with wall thicknesses of $s_0 = 1.5$ mm to $s_0 = 6.5$ mm were used \cite{Alb90}. Albert's empirical formula is \cite{Alb90}:

\begin{equation}
s_1 = -0.741 + 2.734 \cdot \varphi + 1.216 \cdot s_0 - 2.394 \cdot 10^{-3} \cdot d_{a0} + 1.336 \cdot 10^{-2} \cdot \alpha.
\end{equation}

Storoschew and Popov established the following formula for determining wall thickness \cite{Haa83}, \cite{Alb90}, \cite{Sto68}:

\begin{equation}
s_1 = s_0 \cdot \sqrt{\frac{d_{a0}}{d_{a1}}},
\end{equation}

and

\begin{equation}
\Delta s = s_1 + s_0.
\end{equation}

This formula was not worked out by testing, but derived mathematically, which is why it does not differentiate between a scope of validity for thin-walled or thick-walled tubes.

\subsection{Results of the Neural Network predictions}

In this section, we present the results of the proposed GNN-based architecture used for wall thickness prediction. To this end, we present results on rollouts of the FEM simulations as well as the wall thickness. We further provide various ablation studies to analyse the novel approach and discuss the computational requirements for both FEM and GNN-based approximation.

\subsubsection{Results on rollouts}

\begin{figure}[h!]
     \centering
     \begin{subfigure}[b]{0.32\textwidth}
         \centering
         \includegraphics[width=\textwidth]{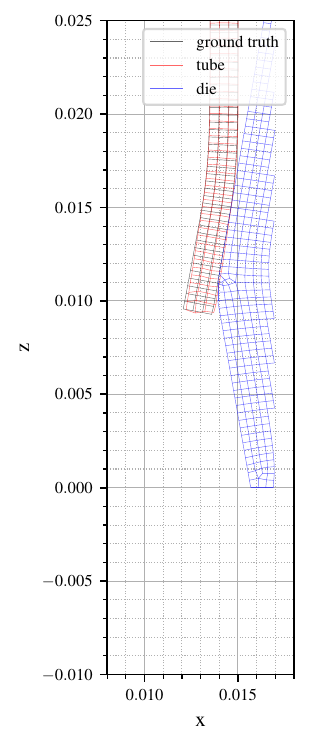}
         \caption{Timestep 500}
     \end{subfigure}
     \hfill
     \begin{subfigure}[b]{0.32\textwidth}
         \centering
         \includegraphics[width=\textwidth]{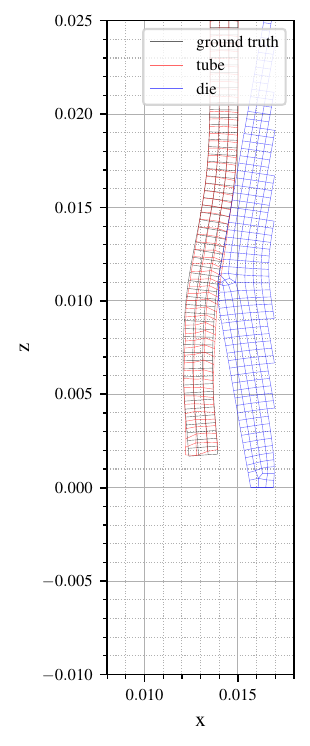}
         \caption{Timestep 1000}
     \end{subfigure}
     \hfill
     \begin{subfigure}[b]{0.32\textwidth}
         \centering
         \includegraphics[width=\textwidth]{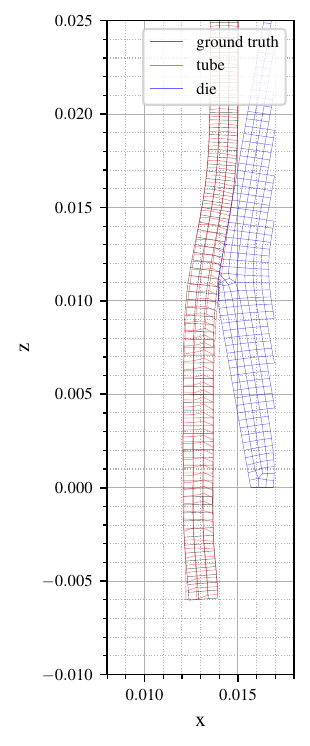}
         \caption{Timestep 1500}
     \end{subfigure}
    \caption{Rollout results for three different timesteps with $\varphi=0.15$, $\alpha=10$}
    \label{fig:rollout_example_diff_timesteps}
\end{figure}

After training our \ac{NN} model as described previously, we test the performance on the unseen parameter configurations as already shown in \autoref{tab:simulation_run_parameters}. Specifically, to generate a sequence we forward integrate using the trained models respective prediction for 4500 timesteps to obtain the tube mesh after the forging process. An exemplary prediction for different timesteps is shown in \autoref{fig:rollout_example_diff_timesteps}.

As can be seen, the predicted meshes match well with the FEM ground truth data indicating a very good representation capability and accuracy of the proposed approach. We note that the prediction tends to deviate at the beginning of the tube during the rollouts, hence where the funnel shape of the collar occurs. However, as the deviation occur only after the tube has left the processing area, these deviations do not harm the models forming predictions in the quasi-stationary operating point which is its main purpose. Furthermore, this does not impose limitations for usage a real-time loop when targeting the quasi-steady operation.
The above results gives us a good intuition about the process and predictions obtained from the rollout, however it is difficult to quantify the general performance of the model as well as the generalization capabilities by visual inspection only. 

Thus, as we are mainly interested in the change of the tube thickness $\Delta s$, we take this as the second important performance metric in addition to the mesh accuracy to rate the models prediction performance. When calculating the respective thickness change at timestep 4500, we obtain the plots shown in \autoref{fig:thickness_change}. Note that we only display the thickness change up to a relative tube length of 0.8 as this is the main processing area we are interested in for downstream tasks like process monitoring and control. After leaving these area, the tube hasn't had any contact with the die.

\begin{figure}[htbp]
     \centering
     \begin{subfigure}{\textwidth}
         \includegraphics[width=\textwidth]{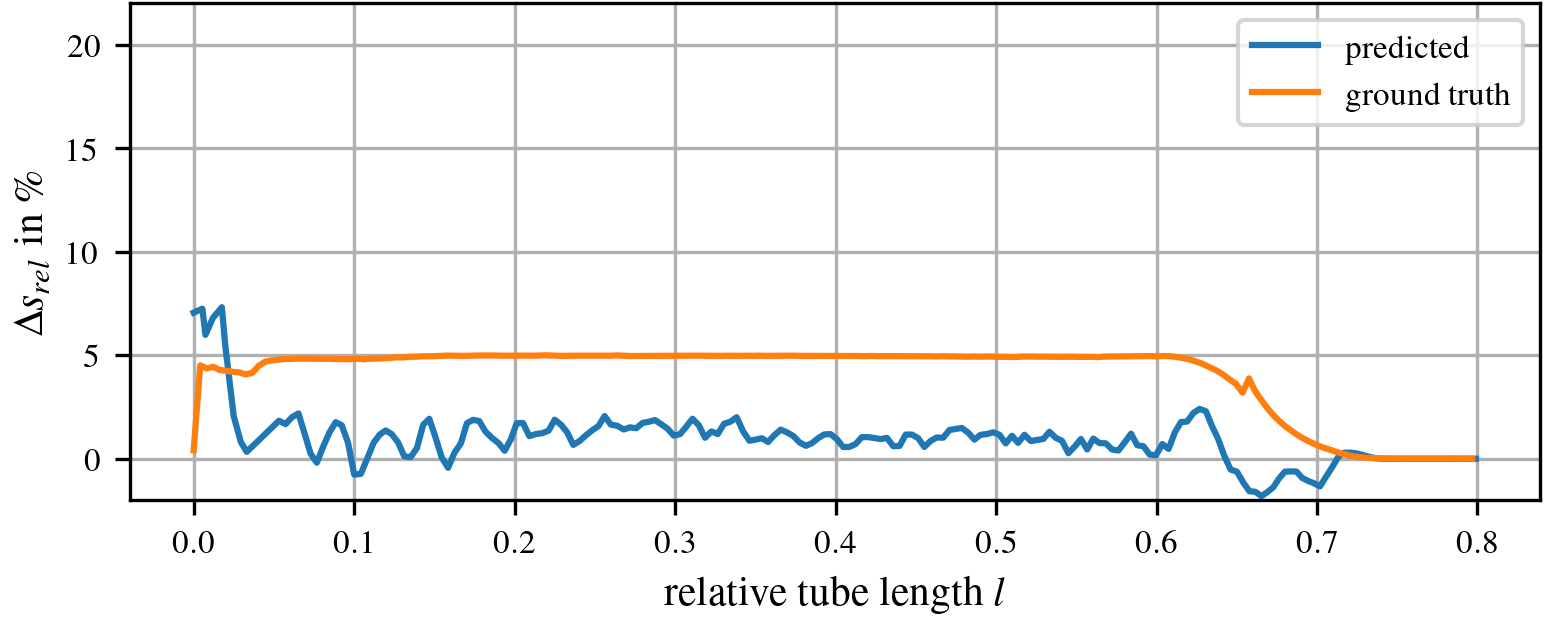}
         \caption{$\varphi = 0.1, \alpha = 17.5$}
     \end{subfigure}
     \begin{subfigure}{\textwidth}
         \includegraphics[width=\textwidth]{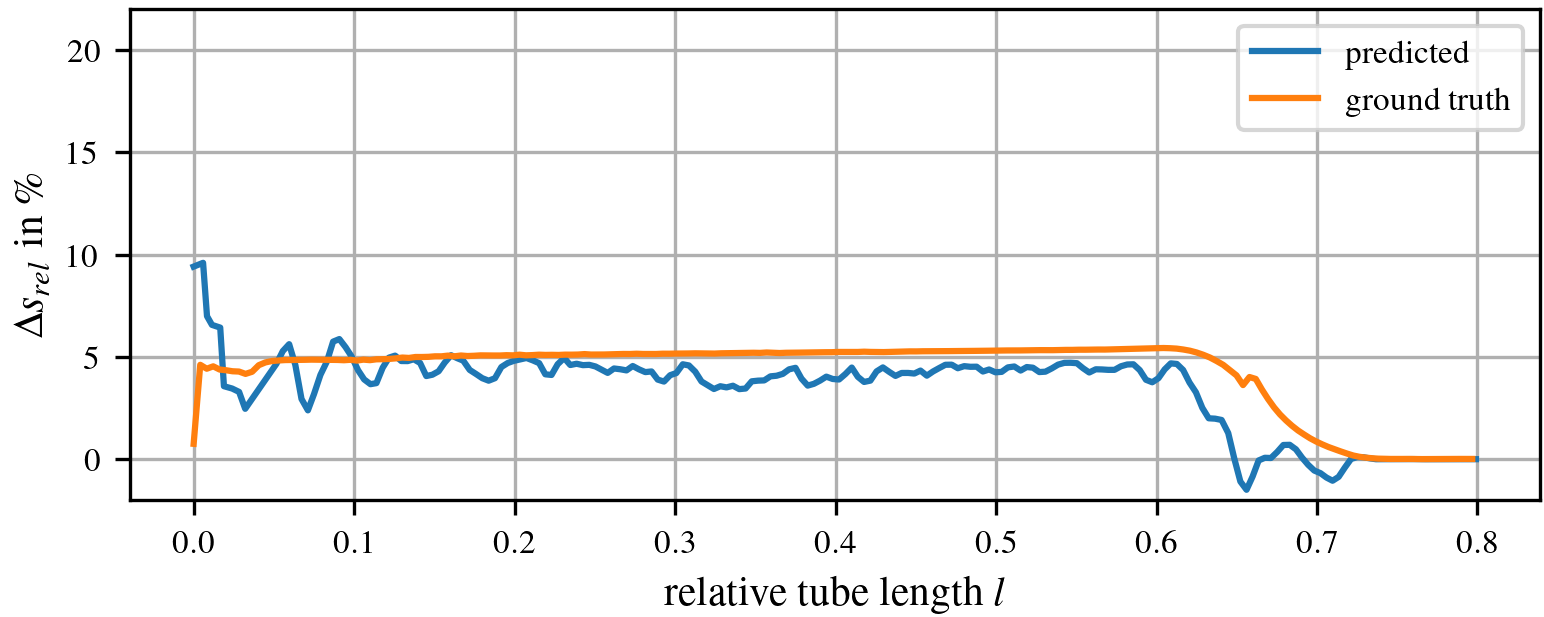}
         \caption{$\varphi = 0.1, \alpha = 22.5$}
     \end{subfigure}
     \begin{subfigure}{\textwidth}
         \includegraphics[width=\textwidth]{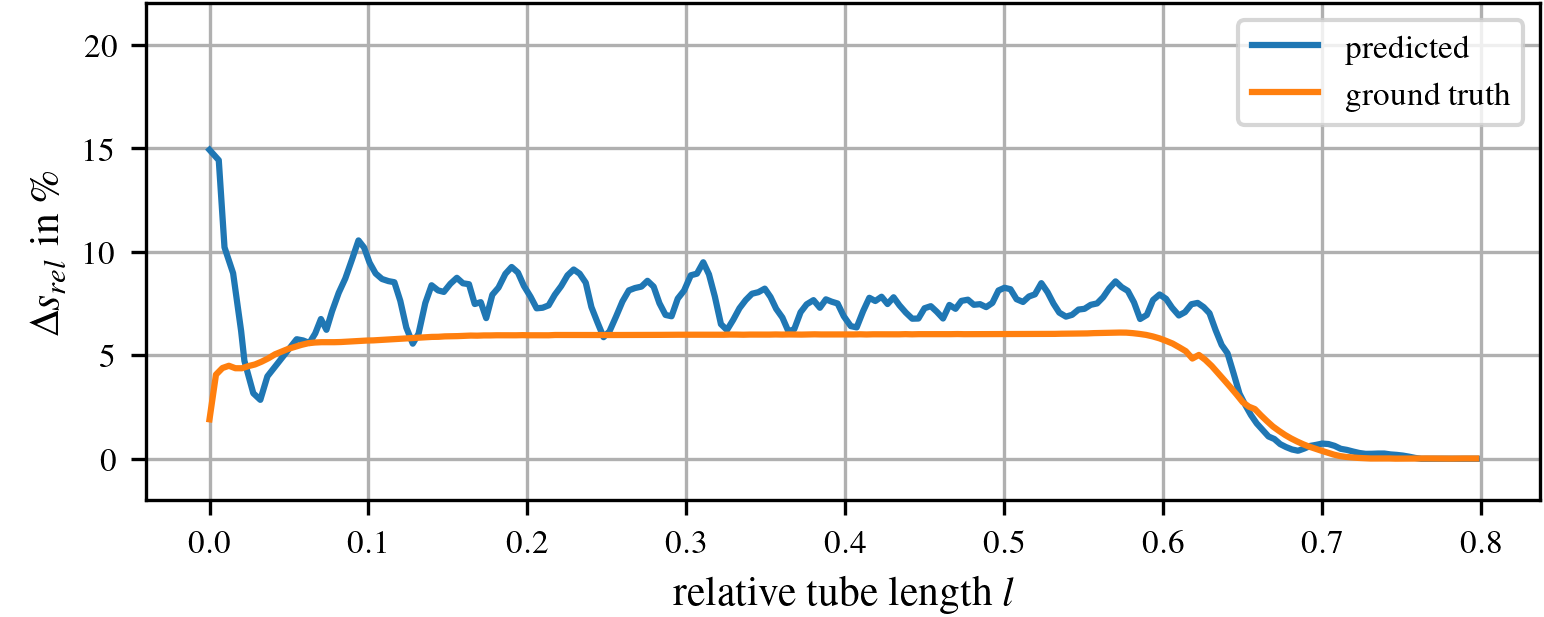}
         \caption{$\varphi = 0.15, \alpha = 10$}
     \end{subfigure}
\end{figure}
\begin{figure}\ContinuedFloat
    \centering
     \begin{subfigure}{\textwidth}
         \includegraphics[width=\textwidth]{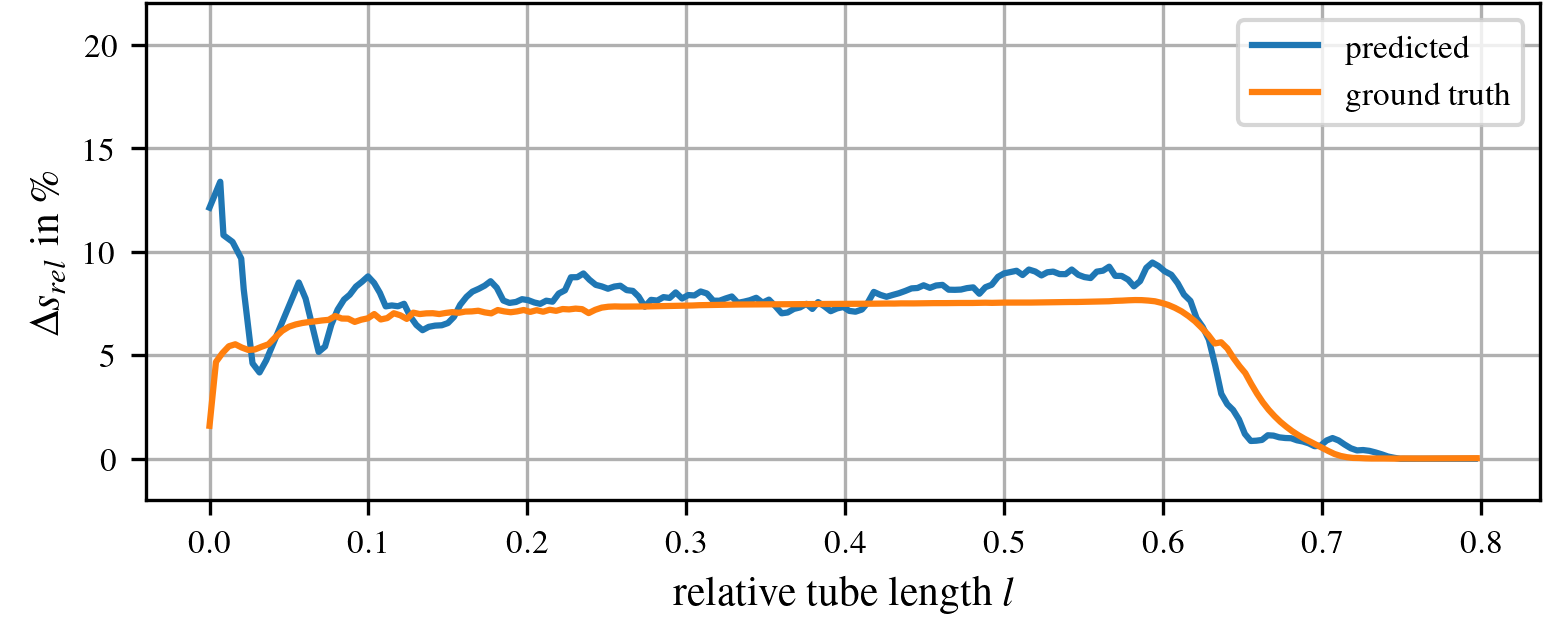}
         \caption{$\varphi = 0.15, \alpha = 15$}
     \end{subfigure}
     \begin{subfigure}{\textwidth}
         \includegraphics[width=\textwidth]{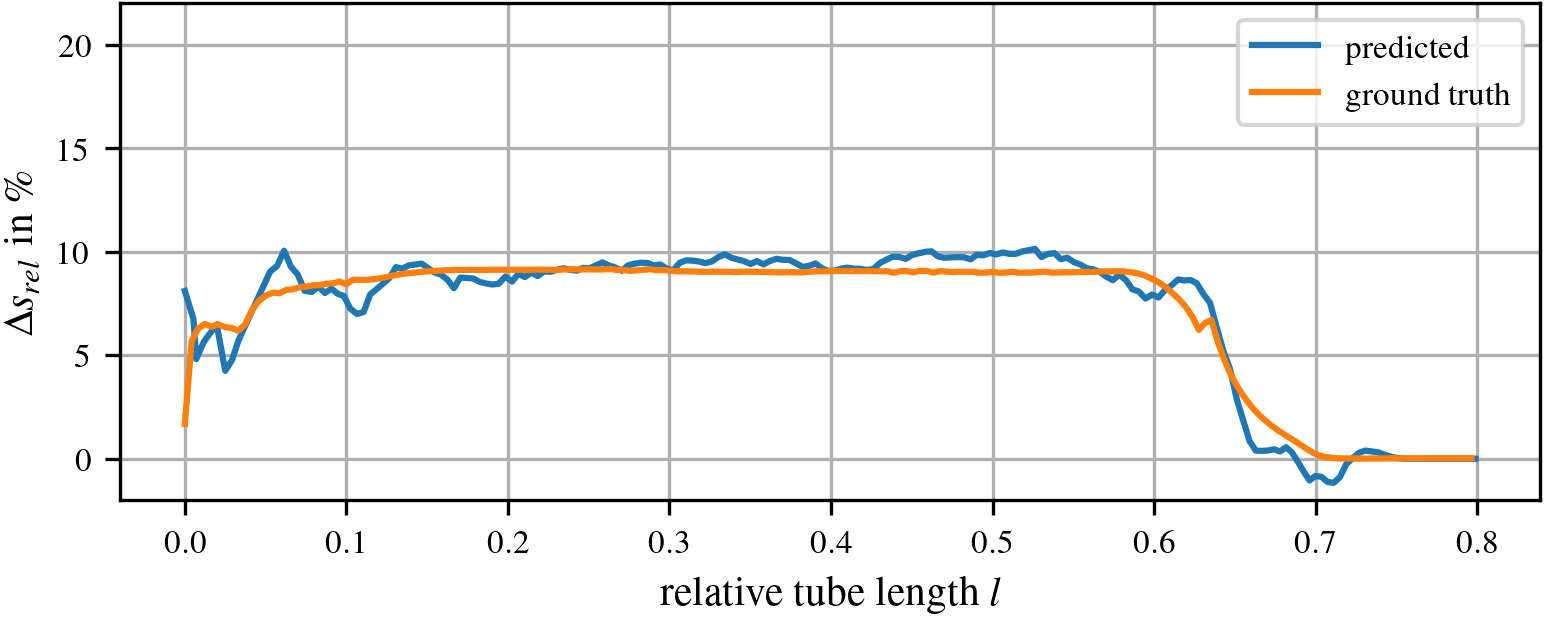}
         \caption{$\varphi = 0.15, \alpha = 20$}
     \end{subfigure}
     \begin{subfigure}{\textwidth}
         \includegraphics[width=\textwidth]{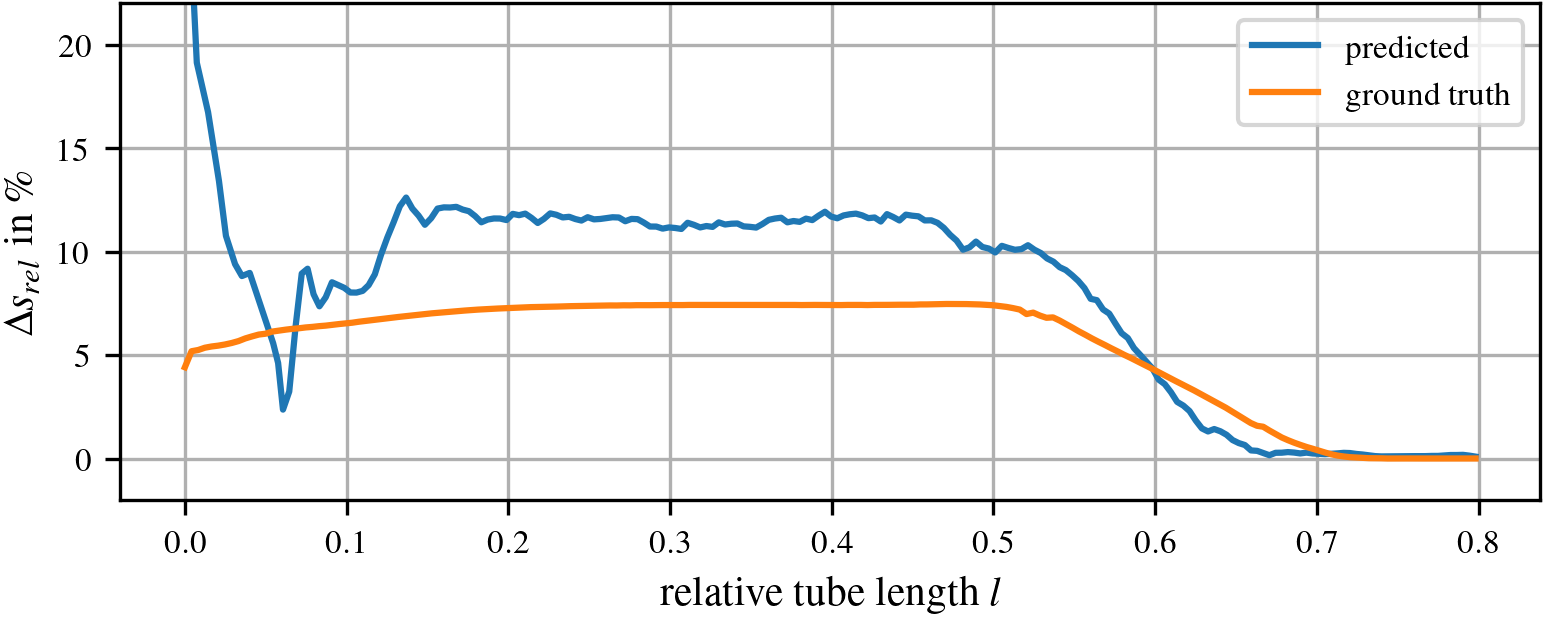}
         \caption{$\varphi = 0.2, \alpha = 5$}
     \end{subfigure}
\end{figure}
\begin{figure}\ContinuedFloat
    \centering
     \begin{subfigure}{\textwidth}
         \includegraphics[width=\textwidth]{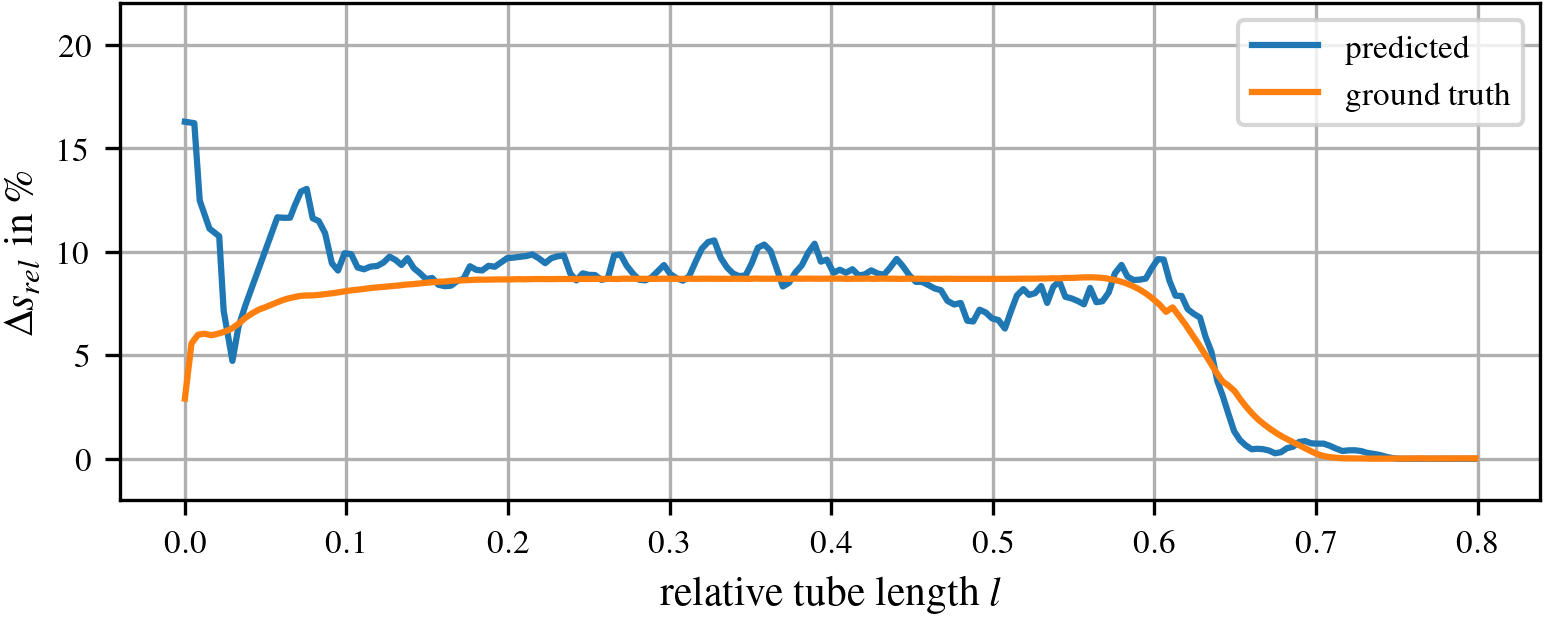}
         \caption{$\varphi = 0.2, \alpha = 12.5$}
     \end{subfigure}
     \begin{subfigure}{\textwidth}
         \includegraphics[width=\textwidth]{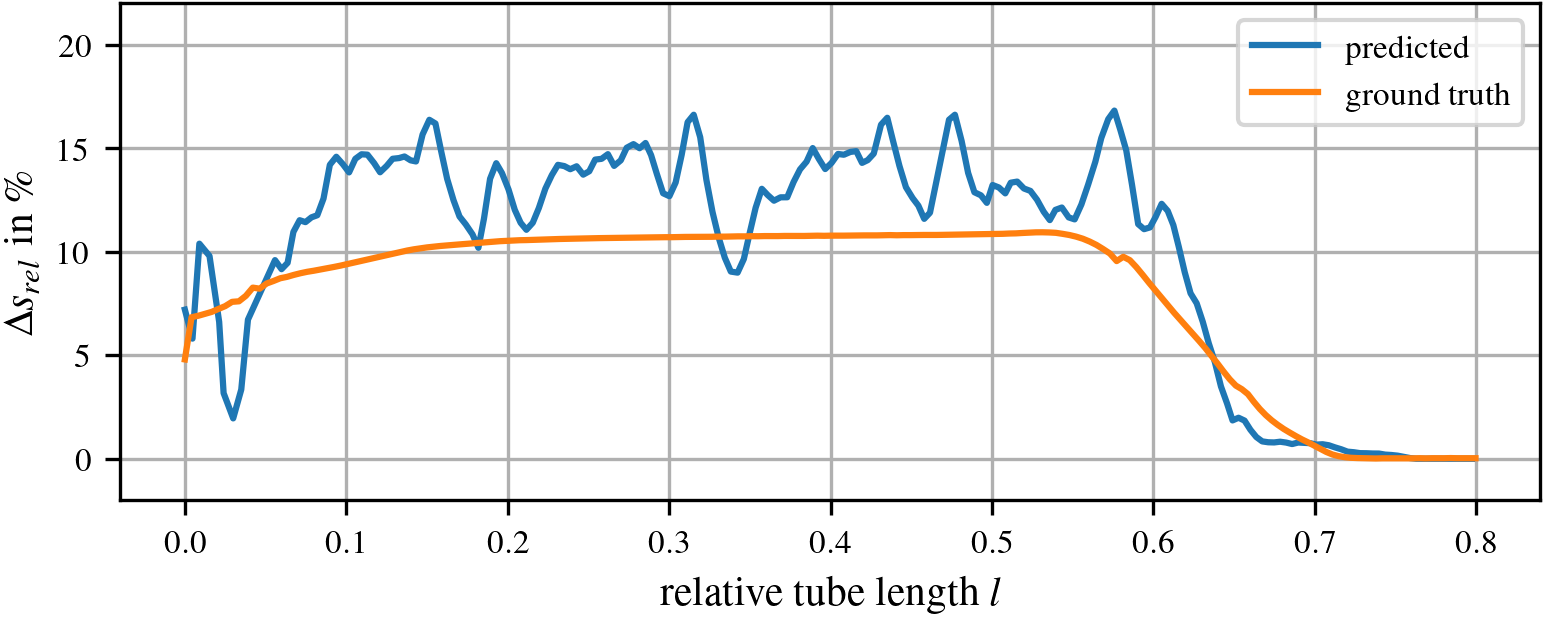}
         \caption{$\varphi = 0.25, \alpha = 10$}
     \end{subfigure}
     \begin{subfigure}{\textwidth}
         \includegraphics[width=\textwidth]{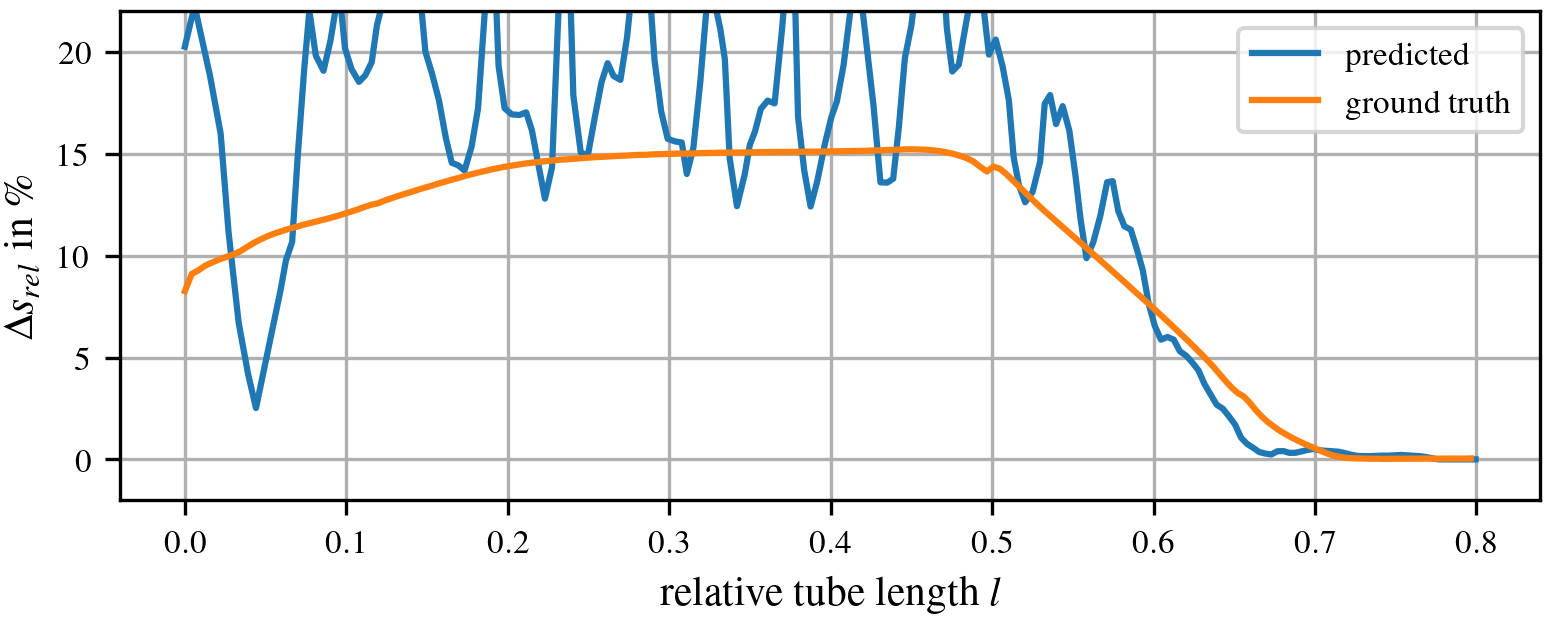}
         \caption{$\varphi = 0.35, \alpha = 7.5$}
     \end{subfigure}
    \caption{Thickness differences $\Delta s$ after deformation for the different test configurations.}
    \label{fig:thickness_change}
\end{figure}

It can be observed that the general trend in the prediction matches the ground truth data quite well, however the prediction appears to be more noisy for high deformation degrees, such as $\varphi=0.25$ and $\varphi=0.35$. Especially the run using $\varphi=0.35$ suffers from very noisy predictions, and generally speaking with the least accurate accuracy. Reasons for this are twofold: on the one hand, this specific run is on the very border of our dataset (see \autoref{tab:simulation_run_parameters}), meaning the density of training data is very limited and this the performance is worse. On the other hand, the general performance of the network is still limited and shows room for further improvements.
We observe again, that the prediction struggles the most at the top of the tube. However, the model is able to recover from initial deviations and produces good predictions after only a few rollout steps. 

\begin{figure}
    \centering
    \includegraphics{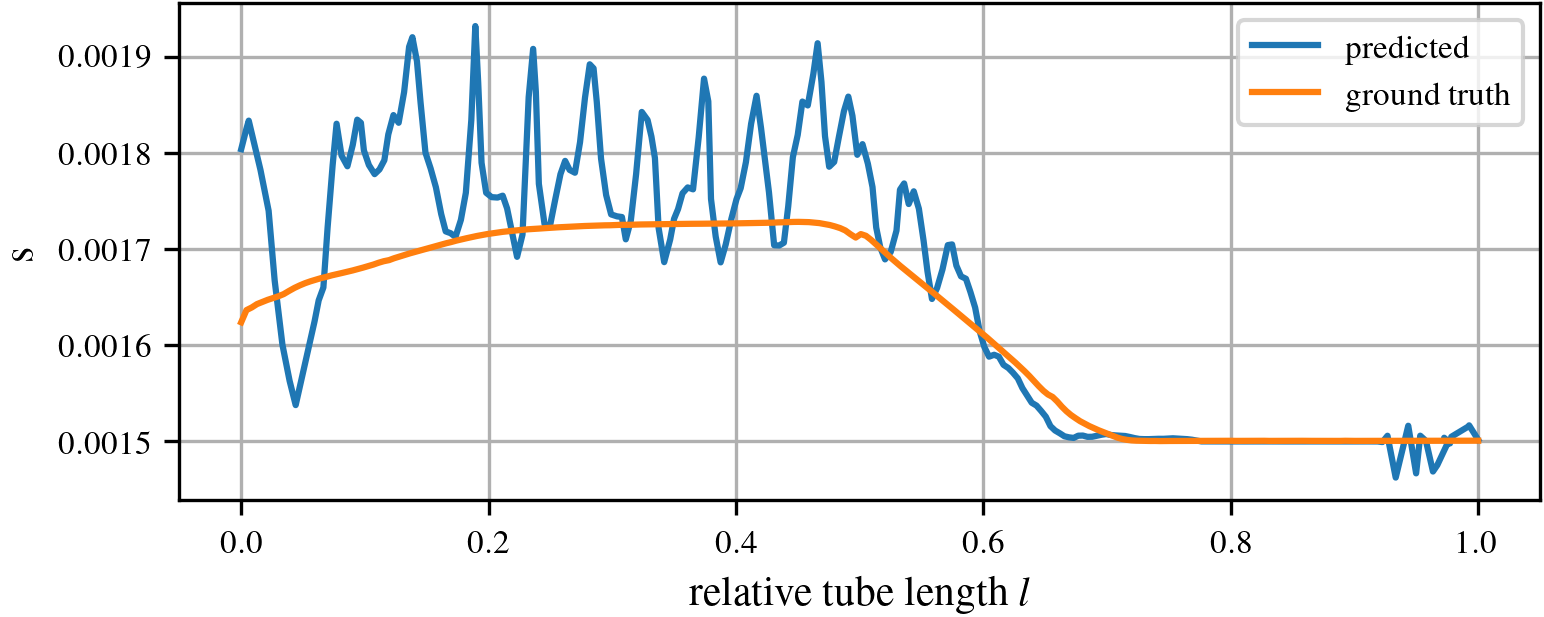}
    \caption{Example of the prediction of the entire tube thickness for $\varphi = 0.35$ and $\alpha = 7.5$}
    \label{fig:example_thickness}
\end{figure}

Furthermore, we show an example of the entire tube thickness in \autoref{fig:example_thickness}.
We can see that at the very end of the tube, where the stamp causes the feed in z-direction, we got undesired thickness changes in comparison to the FEM results. Even though it does not affect us when evaluating the thickness change of the tube after forging, it is something we aim to improve in further refinements of the modeling.

\begin{figure}
    \centering
    \includegraphics{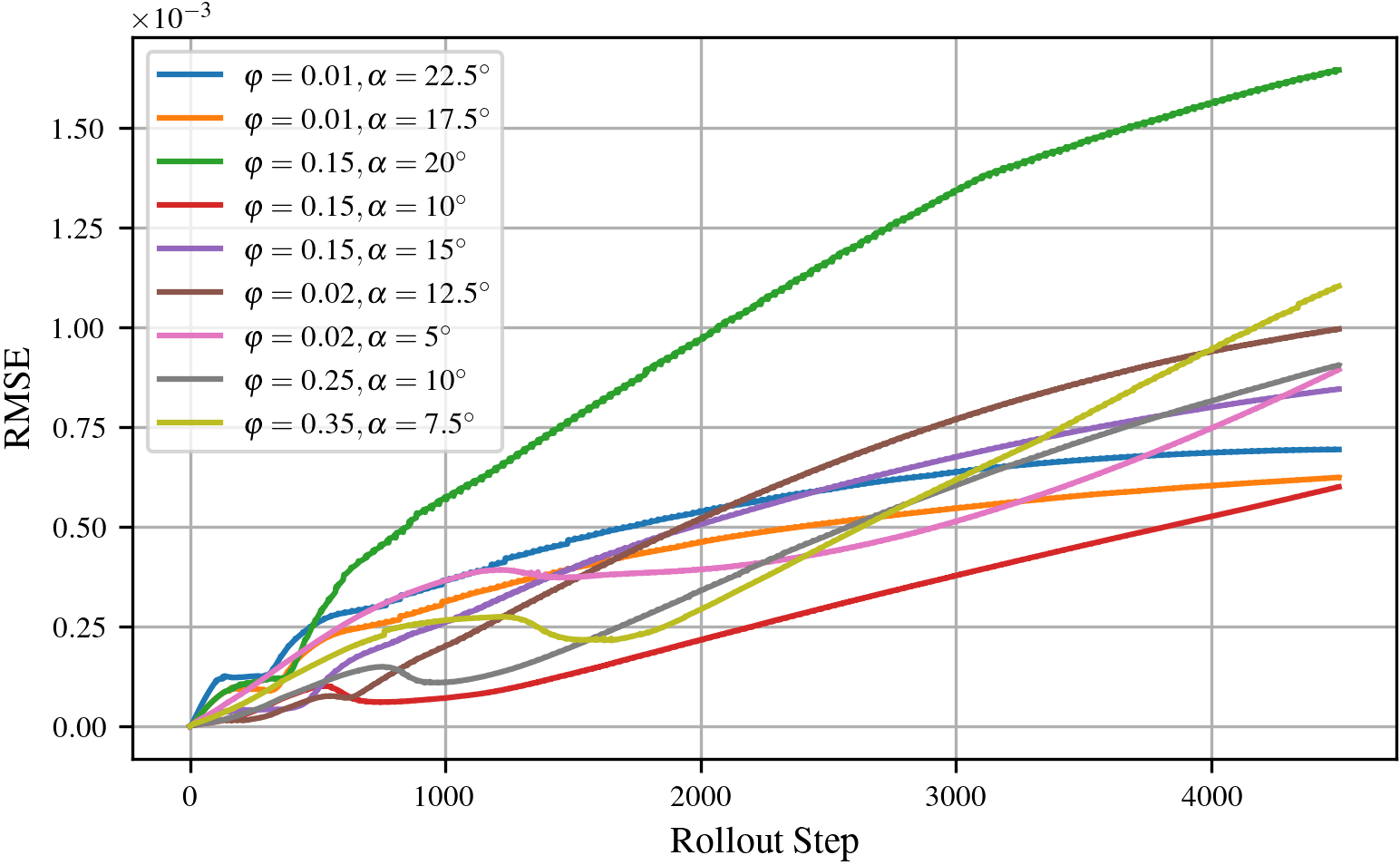}
    \caption{RMSEs for the nine test cases.}
    \label{fig:rollout_rmses}
\end{figure}

Additionally, the rollouts yielded the RMSEs as shown in \autoref{fig:rollout_rmses}.
As we can see, the rollout RMSEs increase over the timesteps which is expected since we use a one-step predictor without any mechanisms to recover from inaccuracies during the sequence prediction. Nevertheless we  don't encounter steep exponential increase either, which would be troublesome for longer rollouts. Instead, as discussed before, the steeper incline is at the initial timesteps where the funnel shape of the collar occurs. After settling from the initial conditions and when we have a stable deformation and thickness change the curve flattens. 

\subsection{Ablation studies}
To increase the understanding of the behaviour of the novel approach, we perform and discuss different ablations in the following. 

\subsubsection{Increasing the step width in comparison to the FEM}
We investigate on the effect when we increase the step size in the \ac{NN} approximation from 1 to 2, 5, 10 and 20, or in other words per simulation step in the \ac{NN}, the distance the stamp moves in z-direction is 2-times, 5-times, 10-times and 20-times as high as in the FEM simulation. Besides that, the training setup remains unchanged. As the metric to compare, we use the area between the thickness curves. This experiment yielded the results presented in \autoref{tab:ablation-step_size}.

\begin{table}[h!]
    \centering
    \caption{ABTC comparison (in mm$^2$) for various step-sizes $step$ and test splits.}
    \label{tab:ablation-step_size}
    \begin{tabular}{l | d{3.2}d{3.2}d{3.2}d{3.2}d{3.2}}
    \toprule
    Configuration & \mc{$step=1$} & \mc{$step=2$} & \mc{$step=5$} & \mc{$step=10$} & \mc{$step=20$} \\
    \midrule
    $\varphi = 0.10, ~\alpha = 17.5^\circ $ & 39.453  & 21.681 & 15.829  & 30.053  & 12.190 \\
    $\varphi = 0.10, ~\alpha = 22.5^\circ$  & 12.691  & 7.452 & 11.651  & 17.559  & 15.808 \\
    $\varphi = 0.15, ~\alpha = 10^\circ$    & 18.337  &  9.453 & 11.786  & 35.454  & 27.181 \\
    $\varphi = 0.15, ~\alpha = 15^\circ$    & 10.922  & 12.259 & 10.138  & 10.328  & 25.218 \\
    $\varphi = 0.15, ~\alpha = 20^\circ$    &  6.774  &  9.642 & 17.971  & 51.214  & 15.225 \\
    $\varphi = 0.20, ~\alpha = 5^\circ$     & 36.060  & 35.813 & 80.439  & 31.876  & 25.682 \\
    $\varphi = 0.20, ~\alpha = 12.5^\circ$  & 12.951  & 37.381 & 36.968  & 34.391  & 102.398 \\
    $\varphi = 0.25, ~\alpha = 10^\circ$    & 29.479  & 43.005 & 38.231  & 45.644  & 35.970 \\
    $\varphi = 0.35, ~\alpha = 7.5^\circ$   & 45.485  & 62.973 & 45.123  & 43.974  & 24.556 \\
    \midrule
    total & 212.152  & 239.658 & 268.135  & 300.494 & 284.227 \\
    \bottomrule
    \end{tabular}
\end{table}

We can observe that step-size two is only marginally worse than step size one.
Smaller step-size tends to result in better performances, which also makes intuitive sense, since the nonlinearities in the nodes' movements are also reduced. With a smaller step-size, the mesh can more closely resemble the fine differences between steps and can basically break down a complex movement into multiple smaller steps. The performances between a step-size of 5 and 20 do not seem to be significantly different, thus, we can conclude that it is well worth varying the step-size in comparison to the FEM to find the optimal one: on the one hand, there is potential for better performance, and on the other hand, and maybe even more important, this allows for faster execution times. We leave a more thorough investigation on this topic and the relationship to other hyperparameter for further research. 

\subsubsection{Simulation without contact between tube and die}

\begin{figure}[h!]
     \centering
     \begin{subfigure}[b]{0.32\textwidth}
         \centering
         \includegraphics[width=\textwidth]{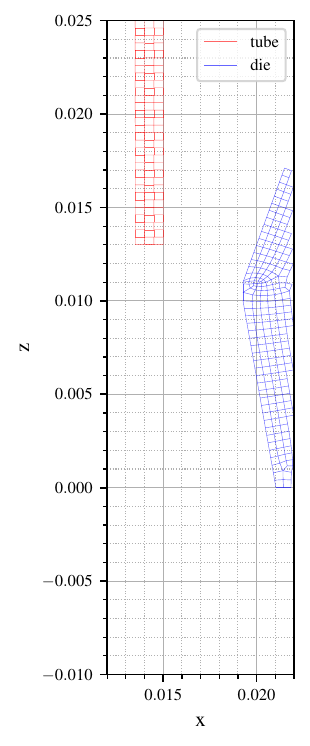}
         \caption{Timestep 0}
     \end{subfigure}
     \hfill
     \begin{subfigure}[b]{0.32\textwidth}
         \centering
         \includegraphics[width=\textwidth]{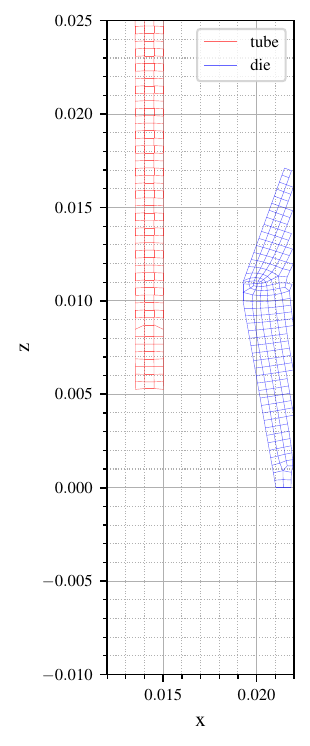}
         \caption{Timestep 500}
     \end{subfigure}
     \hfill
     \begin{subfigure}[b]{0.32\textwidth}
         \centering
         \includegraphics[width=\textwidth]{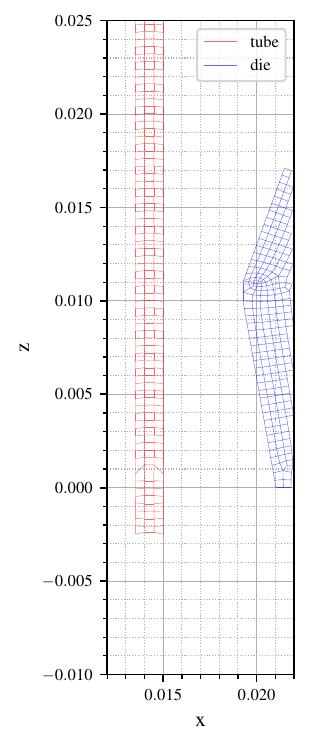}
         \caption{Timestep 1000}
     \end{subfigure}
    \caption{Rollout results for three different timesteps with $\varphi=0.1$, $\alpha=17.5$ when there is no contact with the die.}
    \label{fig:rollout_wo_matrix}
\end{figure}

We provide a further simulation to show the models capability to actually learn to predict forming based on contact with the die. To this end, we perform a simulation where we move the die outwards out of the effective area, such that die and tube are not in contact during operation. Consequently, no dynamic edges between the two objects (compare Figure \ref{fig:contact_transforms_visu}) are created by the framework preventing the message passing between the two objects. We take the same dataset and trajectory as in \autoref{fig:rollout_example_diff_timesteps}, yielding the rollouts as shown in \autoref{fig:rollout_wo_matrix}.
As we can see, no forming of the tube takes place, i.e. it is just pushed down by the movement of the stamp. We can conclude, that the model effectively learns the interaction characteristics between the tube and the die as intended, i.e. it does not simply learn to shift the tube nodes in x-direction based on certain z-coordinate values or hallucinate contact forces.

\subsubsection{Rollout for the whole Dataset}

\begin{figure}[htb]
    \centering
    \includegraphics[width=0.75\textwidth]{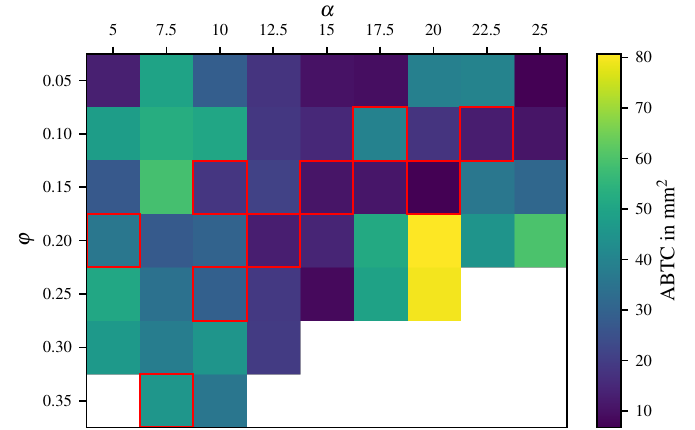}
    \caption{Areas between the thickness curves for the whole dataset (the lower the better). Test cases are highlighted with a red frame.}
    \label{fig:area_thickness_curves}
\end{figure}

To further increase the understanding of the model behavior, we conduct rollouts for all items in the dataset on both train as well as test data. This enables us to understand what the model has learned and in which parameter setting the rollout over multiple timesteps is particularly difficult. 
Even though the model is trained on the majority of the data here, this still gives valuable information as we only train a one-step predictor, but do a rollout over multiple thousand timesteps.
As a measure, we again use the ABTC. The obtained results are shown in \autoref{fig:area_thickness_curves}.

It is evident that the best model prediction performance can be achieved for higher die opening angles $\alpha$ and medium degree of deformation $\varphi$. Hence, for parameter configurations in the middle of the considered parameter range, the model predictions are very good with a low ABTC. We further emphasize that there is no major difference between training and test cases, indicating the good generalization capabilities which are of particular importance from a practical application point of view.

Especially for high deformation degrees, i.e. higher $\varphi$ and $\alpha$, the results in terms of ABTC deteriorates. We have two explanations for this observation: First, medium degrees of deformation lie in the middle of the dataset distribution while we approach the process limits for higher $\varphi$ and $\alpha$. Second, for small $\alpha$, there is way more contact time between the tube and the tool. This yields more dynamic edges between the tube and the tool, and consequently more interaction when transitioning in the z-direction, which makes these scenarios more prone to error propagation when there are slight inaccuracies within the prediction per timestep. We will elaborate more on this in future work.

\subsection{Execution times}

One of the main benefits of the \ac{NN} approach is the drastic reduction in computation time during deployment. To showcase this fact, we ran a test where we measured the time needed for 4500 simulation steps on a CPU as well as a GPU. We used the test dataset with $\varphi = 0.20, ~\alpha = 5$ as this is expected to be the slowest due to the flattest angle of $5\degree$ and therefore the most dynamically added edges during model calculation. This yielded the times shown in \autoref{tab:execution_times}. The CPU used was a Intel Core i7-11700 and the GPU a NVIDIA Quadro RTX 5000. The used model consisted of 2.88 million trainable parameters. The CPU used for the calculation of the \ac{FEM} simulations was a Intel Core i9-7980XE CPU and a NVIDIA GeForce GTX 1070 GPU. Even though these systems use different hardware, the general different is still big enough to guarantee a significant speed increase of the \ac{GNN} model.

\begin{table}[h!]
    \caption{Comparison of execution times.}
    \label{tab:execution_times}
    \centering
    \begin{tabular}{c | c | c }
    \toprule
    Approach & Device & Time \\
    \midrule
    FEM & CPU & $\sim$ 301 minutes \\
    NN & CPU & $\sim$ 23 minutes \\
    NN & GPU & $\sim$ 3.5 minutes \\
    \bottomrule
    \end{tabular}
\end{table}

\section{Conclusion}
\label{sec:conclusion}

We presented a novel approach for predicting wall thickness changes in tubes during the nosing forging process. We provided a thorough analysis of the parameters which influence the nosing process and based on this information, we set-up an FEM simulation of the nosing process. Below are the main contributions of this work:

\begin{itemize}
    \item A thorough analysis of the nosing process and influencing parameters
    \item We developed a detailed FEM simulation to analyze effects of varying process parameters in nosing.
    \item We developed a surrogate model for real-time predictions based on \ac{GNN}s.
    \item We extended existing \ac{NN} architectures to incorporate the available information of the nosing process in the architecture.
    \item We proposed a novel performance metric to evaluate results of nosing processes, the \ac{ABTC}.
\end{itemize}  

The separation of edge types into their corresponding encoders resulted in more accurate models and opens up the possibility of using accurate surrogate models in a closed loop production process further down the line.

To enhance the applicability of the model, further research is required. Extending the model to encompass a variety of material parameters, embedded in node or edge information, and a more encompassing parameter space, such as initial wall thicknesses and different lubrications, will increase its versatility and applicability. Using multiple encoders as part of the integration of knowledge about the physical process also needs further investigation, especially the capability of disentangling the motions in different dimensions, especially with regards to radial- and tangential forces. Furthermore, we plan to compare the FEM data with a real experimental setup and trials in future work, as well as to integrate the model into real-time control systems of production processes

\backmatter
\section*{Statements and Declarations }
\bmhead{Funding} 
This research did not receive any specific grant from funding agencies in the public, commercial, or not-for-profit sectors.

\bmhead{Conflict of interest/Competing interests}
There are no conflicts of interest to declare.

\bmhead{Ethics approval and consent to participate}
Not applicable

\bmhead{Consent for publication}
Not applicable

\bmhead{Data availability}
Not applicable

\bmhead{Materials availability}
Not applicable

\bmhead{Code availability}
Not applicable

\bmhead{Author contribution}
The individual contrubition of authors is equivalent.

\bmhead{Declaration of Generative AI and AI-assisted technologies in the writing process}

During the preparation of this work the authors used ChatGPT4 in order to support text formulation/translation. After using this tool, the authors reviewed and edited the content as needed and take full responsibility for the content of the publication.

\bibliography{sources}
\end{document}